\documentclass{EngC}

\usepackage[rightcaption,raggedright]{sidecap}
\usepackage{framed}         
\usepackage{soul}           
\usepackage{bm}

\usepackage[agsm]{harvard}
\usepackage{comment}
\usepackage{booktabs,tabularx}

\usepackage{rotating}
\usepackage{floatpag}
\rotfloatpagestyle{empty}

\usepackage{amsmath}
\usepackage{amsthm}
\usepackage{amsfonts}
\usepackage{graphicx}

\usepackage{multind}
\makeindex{authors}
\makeindex{subject}

\theoremstyle{plain}

\theoremstyle{definition}

\theoremstyle{remark}

\def\bx{\ensuremath{{\bf x}}}
\def\br{\ensuremath{{\bf r}}}

\def\balpha{\ensuremath{\bm{\alpha}}}

\def\by{\ensuremath{{\bf y}}}
\def\bz{\ensuremath{{\bf z}}}

\def\bh{\ensuremath{{\bf h}}}

\def\bW{\ensuremath{{\bf W}}}
\def\bA{\ensuremath{{\bf A}}}

\def\bk{\ensuremath{{\bf k}}}
\def\bn{\ensuremath{{\bf n}}}

\def\bP{\ensuremath{{\bf P}}}
\def\bQ{\ensuremath{{\bf Q}}}

\def\bF{\ensuremath{{\bf F}}}

\def\bD{\ensuremath{{\bf D}}}
\def\bB{\ensuremath{{\bf B}}}
\def\bI{\ensuremath{{\bf I}}}
\def\bL{\ensuremath{{\bf L}}}
\def\bS{\ensuremath{{\bf S}}}
\def\bX{\ensuremath{{\bf X}}}
\def\bZ{\ensuremath{{\bf Z}}}
\def\bH{\ensuremath{{\bf H}}}
\def\bN{\ensuremath{{\bf N}}}

\def\cP{\ensuremath{{\mathcal P}}}
\def\cT{\ensuremath{{\mathcal T}}}
\def\cF{\ensuremath{{\mathcal F}}}
\def\cA{\ensuremath{{\mathcal A}}}
\def\cR{\ensuremath{{\mathcal R}}}
\def\cM{\ensuremath{{\mathcal M}}}

\def\cS{\ensuremath{{\mathcal S}}}
\def\cQ{\ensuremath{{\mathcal Q}}}
\def\cH{\ensuremath{{\mathcal H}}}

\def\cU{\ensuremath{{\mathcal U}}}
\def\cD{\ensuremath{{\mathcal D}}}
\def\cC{\ensuremath{{\mathcal C}}}
\def\cG{\ensuremath{{\mathcal G}}}
\def\cK{\ensuremath{{\mathcal K}}}
\def\cL{\ensuremath{{\mathcal L}}}
\def\cJ{\ensuremath{{\mathcal J}}}

\begin{document}
\title{Deep Algorithm Unrolling for Biomedical Imaging}

\author{Yuelong Li, Or Bar-Shira, Vishal Monga and Yonina C. Eldar}

\maketitle

\chapter{Deep Algorithm Unrolling for Biomedical Imaging}\label{chap:deep_unrolling}

\author{Yuelong Li, Or Bar-Shira, Vishal Monga and Yonina C. Eldar}

\section{Introduction}\label{sec:unroll_intro}
Model-based inversion has played a dominant role in biomedical imaging prior to
deep learning gaining widespread popularity and broad recognition. Model-based
techniques rely on a \emph{forward model} derived by modeling the imaging
process analytically based on physical laws. Typically, the forward model is
formulated as:
\[
	\by=\cF(\bx)+\bn,
\]
where $\by\in\mathbb{R}^m$ is the observation, $\bx\in\mathbb{R}^n$ is
the underlying image data to be recovered, $\cF$ is the mapping from image
domain to observations, and $\bn\in\mathbb{R}^m$ is the corruptive noise
process.

Under many circumstances, the problem of estimating $\bx$ from $\by$
is ill-posed, because the imaging process is often compressive
due to practical constraints~\cite{eldar2012compressed}. For instance, in Medical Resonance Imaging (MRI)
one may undersample the image in order to accelerate the
acquisition process; in Computed Tomography (CT) a reduction of dose through sparse views in
X-ray radiation is generally preferred for patients' safety which translates
into an underdetermined forward model. Therefore, to reliably estimate $\bx$,
a prior structure is often incorporated
either by capturing physical principles or through manual
handcrafting. The underlying image $\bx$ can then be
estimated by solving a regularized optimization problem:
\begin{equation}
	\min_{\bx}\rho(\by,\cF(\bx)) + \lambda\cR(\bx),\label{eq:unroll_reg_inv}
\end{equation}
where $\rho$ is a metric measuring the deviation in the measurements, $\cR$ is a regularization
functional capturing prior structure, and $\lambda$ is a regularization
parameter controlling the regularization strength.
Problem~\eqref{eq:unroll_reg_inv} is generally solved via an iterative
optimization algorithm.

In contrast to the model-based framework, $\bx$ can alternatively be estimated
by learning a regression mapping from $\by$ (or its transformed version such as compressive samples or Fourier coefficients) to
$\bx$ through a data-driven approach. The regression mapping can be chosen from
various machine learning models, among which Deep Neural Networks (DNN) are
increasingly popular nowadays. In particular, a purely data-driven approach
adopts a generic DNN without incorporating any physical laws or prior
structure. Instead, it relies on abundant training data to learn a huge number
of network parameters, and in turn the form of the regression mapping. When the
training data is adequate, this approach can be highly successful because DNNs
are extremely expressive and can learn to adapt towards complicated mappings
which are difficult to characterize and design manually. In addition, with the
aid of highly optimized deep learning platforms and hardware accelerators such
as Graphics Processing Units (GPU), inference via DNN can be performed rapidly,
which gives rise to fast execution speed in practice. This is in contrast with
traditional iterative algorithms which need to repeat certain operations a
large number of times sequentially, and can be quite slow comparatively.

Nevertheless, modern DNNs typically carry a deep hierarchical architecture
composed of many layers and network parameters (can be millions) and are thus
often difficult to interpret. It is hard to discover what is the exact form that is
learned by the DNNs and what are the roles of the individual parameters. In other words, DNNs, as an intact machine learning system, generally lacks the ability for humans to disintegrate, analyze and draw analogies. The ability to understand the sequence of operations of a trained learning system that allows the manipulation of its inner mechanism in a principled, predictable fashion is commonly referred to as \emph{interpretability}. Lack
of interpretability is an important concern as it is usually the key to
conceptual understanding and advancing knowledge frontiers. In addition, in
practical biomedical applications, interpretability is key to ensure trust
in the reconstruction algorithm, as experts need to have proper understanding
of the origin of the artifacts introduced by the algorithm and identify
potential failure cases~\cite{min2017deep}. In contrast, traditional iterative
algorithms are usually highly interpretable because they are developed via
modeling underlying physical processes and/or capturing prior domain knowledge. 

In addition to interpretability, generalizability is another fundamental
requirement in biomedical applications. It is well known that generic DNNs
heavily rely on the quantity and quality of training data in order to achieve
empirically superior performance. In other words, when training data is
deficient, the issue of over-fitting may manifest itself in the form of
significantly degraded network performance. Indeed, this issue can be so
serious that the DNNs may underperform traditional iterative algorithms.
This phenomenon
is especially apparent for DNNs with high model capacity, such as very deep or
ultra wide neural networks. Unlike natural images, in biomedical imaging the
data collection is an expensive and time-consuming procedure, and thus
generalizability is especially important.

\begin{figure*}
	\includegraphics[width=\textwidth]{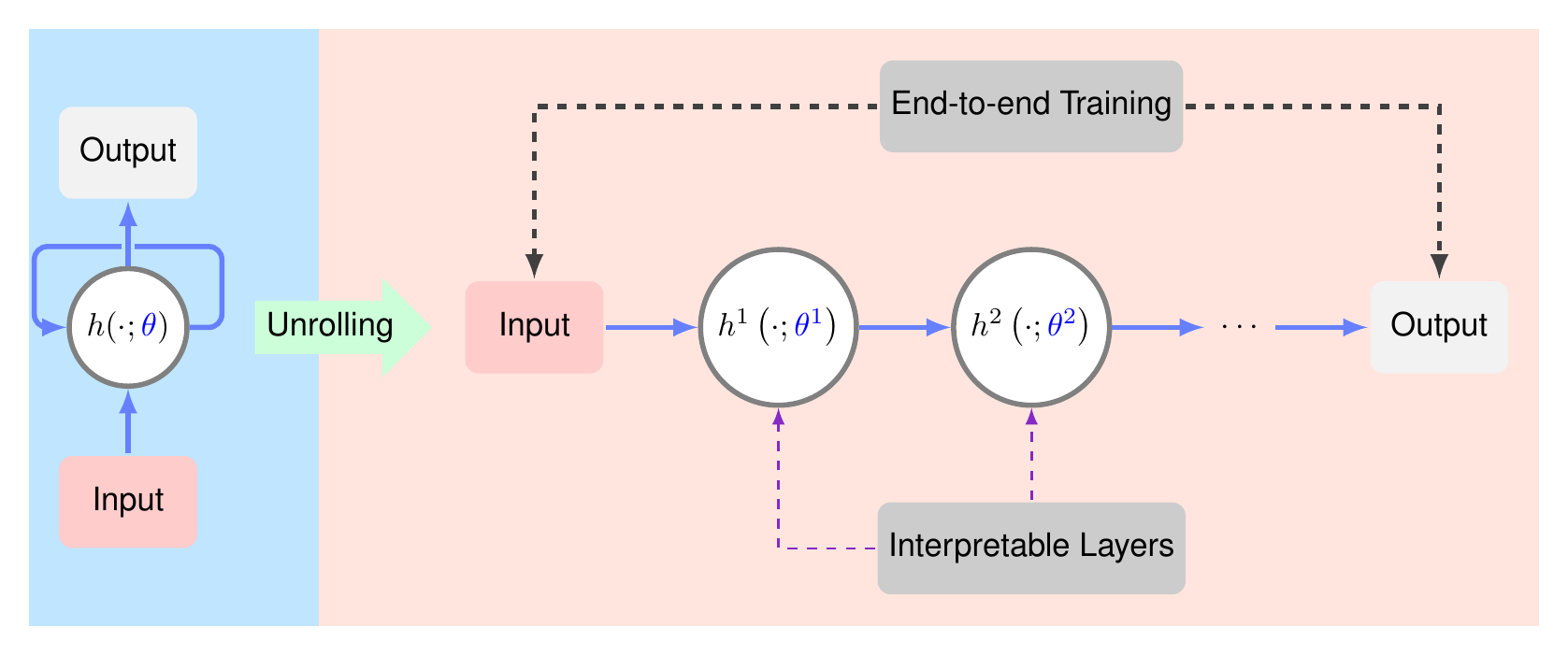}
	\caption{A high-level overview of algorithm unrolling: given an iterative algorithm (left), a corresponding deep network (right) can be generated by cascading its iterations $h$. The iteration step $h$ (left) is executed a number of times, resulting in the network layers $h^1, h^2, \dots$ (right). Each iteration $h$ depends on algorithm parameters $\theta$, which are transferred into network parameters $\theta^1,\theta^2,\dots$ and learned from training datasets through end-to-end training. In this way, the resulting network could achieve better performance than the original iterative algorithm. In addition, the network layers naturally inherit interpretability from the iteration procedure. The learnable parameters are colored in blue.}\label{fig:summary}
\end{figure*}

In the seminal work of Gregor and LeCun~\cite{gregor_learning_2010}, a
promising technique called algorithm unrolling was developed that has helped
connect iterative algorithms such as sparse coding techniques to DNNs. This work has inspired a growing list of follow-ups in different fields of biomedical research: Compressive Sensing (CS)~\cite{yang2018admm}, Computed Tomography (CT)~\cite{hauptmann2018model}, Ultrasound~\cite{solomon_deep_2018,mischi2020deep}, to name a few.
Figure~\ref{fig:summary} provides a high-level illustration of this framework.
Starting with an iterative procedure, each iteration of the algorithm is represented as one layer of a
network. Concatenating these layers forms a deep neural network whose architecture borrows from the optimization method.  Passing
through the network is equivalent to executing the iterative algorithm a finite
number of times. In addition, the algorithm parameters (such as the model parameters and regularization coefficients) transfer to the network parameters. The network may be trained using back-propagation resulting in model parameters that are learned from real world training sets. In this way, the
trained network can be naturally interpreted as a parameter optimized algorithm,
effectively overcoming the lack of interpretability in most conventional neural
networks~\cite{monga2019algorithm}.

Traditional iterative algorithms generally entail
significantly fewer parameters compared with popular neural networks.
Therefore, the unrolled networks are highly parameter efficient and require
less training data. In addition, unrolled networks naturally inherit prior
structure and domain knowledge rather than learn them from intensive training
data. Consequently, they tend to generalize better than generic networks, and
can be computationally faster as long as each algorithmic iteration (or the
corresponding layer) is not overly expensive.

In this chapter, we review biomedical applications and breakthroughs via
leveraging algorithm unrolling,  an important technique that bridges between
traditional iterative algorithms and modern deep learning techniques. To
provide context, we start by tracing the origin of algorithm unrolling and
providing a comprehensive tutorial on how to unroll iterative algorithms into
deep networks. We then extensively cover algorithm unrolling in a wide variety
of biomedical imaging modalities and delve into several representative
recent works in detail. Indeed, there is a rich history of iterative algorithms
for biomedical image synthesis, which makes the field ripe for unrolling
techniques. In addition, we put algorithm unrolling into a broad perspective,
in order to understand why it is particularly effective and discuss recent
trends. Finally, we conclude the chapter by discussing open challenges, and
suggesting future research directions.

\section{Development of Algorithm Unrolling}\label{sec:unroll_development}
This section introduces algorithm unrolling. We begin by reviewing Learned
Iterative Shrinkage and Thresholding Algorithm (LISTA), the first work that
employs the algorithm unrolling strategy. To this end, we review the classical
Iterative Shrinkage and Thresholding Algorithm (ISTA)
(Section~\ref{ssec:unroll_ista}), and then discuss how it can be unrolled into
a deep network (Section~\ref{ssec:unroll_lista}). Next we consider several
related theoretical studies to gain a deeper understanding of LISTA
(Section~\ref{ssec:unroll_theory}). We then introduce the general idea of
algorithm unrolling in an abstract fashion and extend it to generic iterative
algorithms (Section~\ref{ssec:unroll_general}).

\subsection{Iterative Shrinkage and Thresholding Algorithm}\label{ssec:unroll_ista}
In many practical problems, the signal of interest cannot be observed directly,
but must be inferred from observable quantities. In the simplest approximation,
which often suffices, there is a linear relationship between the signal of
interest and the observed quantities~\cite{daubechies2004iterative}.  If we
model the signal of interest by a vector $\bx$, and the derived quantities by
another vector $\by$, we can cast the problem of inferring $\bx$ from $\by$ as
a linear inverse problem. A basic linear inverse problem admits the following
form:
\begin{equation}
    \by=\bA\bx+\bn,\label{eq:unroll_basic_model}
\end{equation}
where $\bA\in\mathbb{R}^{n\times m}$  and $\by\in\mathbb{R}^n$ are known, $\bn$
is an unknown noise vector, and $\bx$ is the unknown signal of interest to be
estimated. 

In practice, it is often the case that $\bA$ has fewer rows than columns
($n<m$) so that recovery of $\bx$ from $\by$ becomes an ill-posed
problem. It is therefore impossible to recover $\bx$ without
introducing additional assumptions on its structure. A popular strategy is to
assume that $\bx$ is sparse or admits a compressible (sparse) representation,
meaning it is sparse in a transformed domain~\cite{eldar2012compressed}.

In sparse coding, we seek a vector $\bx\in\mathbb{R}^m$ such that $\by \approx \bW \bx$
while encouraging as many coefficients in $\bx$ to be zero (or small in
magnitude) as possible~\cite{chen2001atomic}. Here $\bW$ is commonly called a \emph{dictionay} whose columns, typically named \emph{atoms}, can either be from a standard basis (or frames) such as Fourier or wavelet basis, or be learned from real data. A popular technique to
recover $\bx$ is by solving an unconstrained convex minimization problem:
\begin{equation}
	\min_{\bx\in\mathbb{R}^m} \frac{1}{2}\left\|\by-\bW \bx\right\|_2^2+\lambda\|\bx\|_1,\label{eq:unroll_l1_min}
\end{equation}
where $\lambda > 0$ is a regularization parameter that controls the sparsity of
the solution. A well known class of methods for solving~\eqref{eq:unroll_l1_min}
are proximal methods such as ISTA~\cite{daubechies2004iterative}, which perform the following iterations:
\begin{equation}
	\bx^{l+1} = \cS_\lambda\left\{\left(\bI-\frac{1}{\mu}\bW^T\bW\right)\bx^l +\frac{1}{\mu}\bW^T\by\right\},\quad l=0,1,\dots.\label{eq:unroll_ista_iter}
\end{equation}
Here, $\bI\in\mathbb{R}^{m\times m}$ is the identity matrix, $\mu$ is a
positive parameter that controls the iteration step size, $\cS_\lambda(\cdot)$
is the soft-thresholding operator defined as
\begin{equation}
	\cS_\lambda(x)=\mathrm{sign}(x)\cdot\max\{|x|-\lambda,0\},\label{eq:unroll_soft_thresh_op}
\end{equation}
for a scalar $x$, and $\cS_\lambda(\cdot)$ operates element-wise on vectors and matrices.

\subsection{LISTA:\@ Learned Iterative Shrinkage and Thresholding Algorithm}\label{ssec:unroll_lista}
The slow convergence rate of ISTA can be problematic in real-time applications.
Furthermore, the matrix $\bW$ may not be known exactly. In their seminal work,
Gregor {\it et al.} propose a highly efficient learning-based method that computes
good approximations of optimal sparse codes in a fixed amount of
time, with the help of $\bW$ learned from real
data~\cite{gregor_learning_2010}.

Specifically, the authors develop deep algorithm unrolling as a strategy for
designing neural networks which integrate domain knowledge. In this approach, the
network architecture is tailored to a specific problem, based on a well-founded
iterative mathematical formulation for solving the problem at hand. This leads
to increased convergence speed and accuracy with respect to the standard
iterative solution, and interpretability and robustness relative to a black-box
large-scale neural network.

\begin{figure*}
	\centering
	\includegraphics[width=\textwidth]{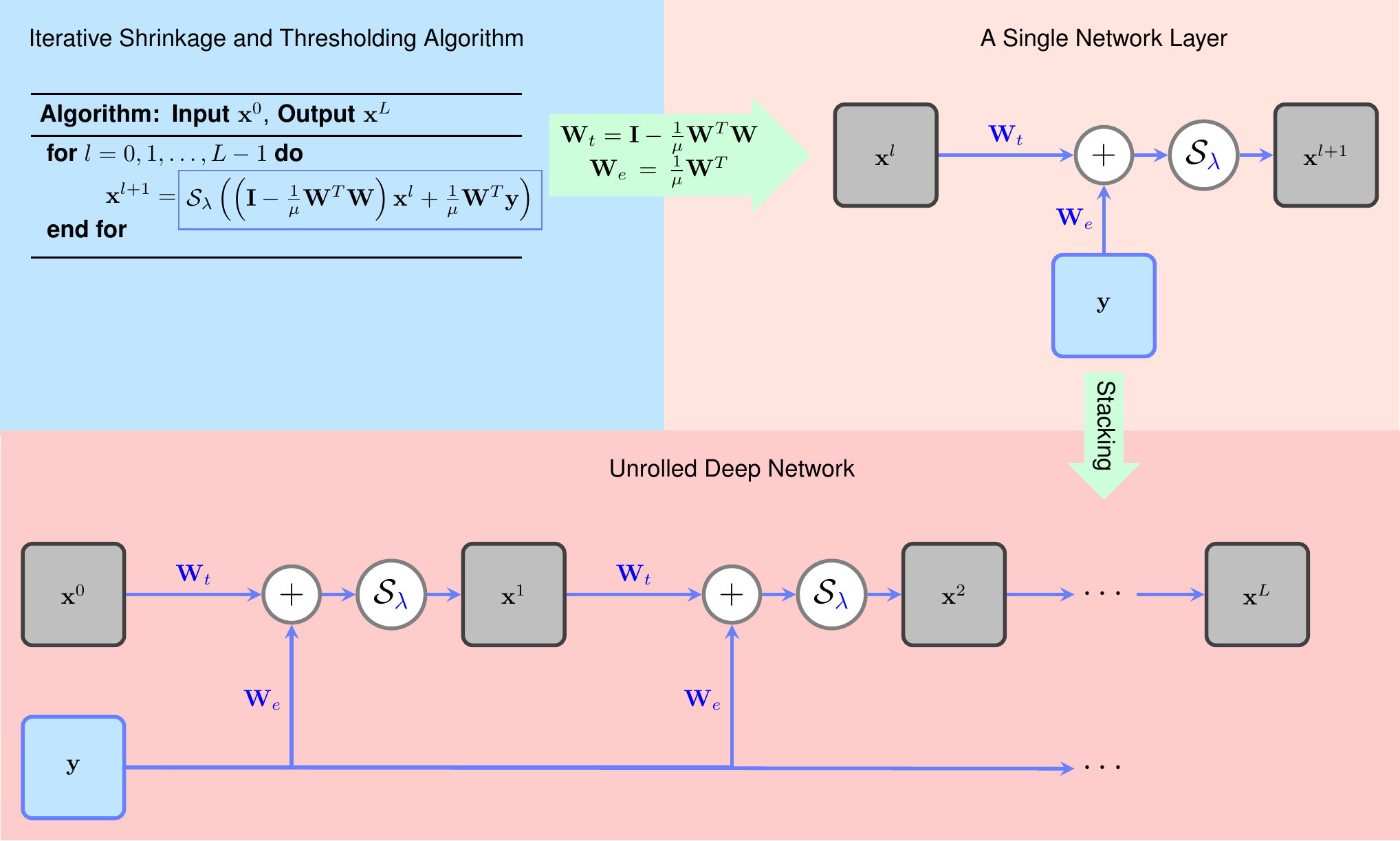}
	\caption{Illustration of LISTA:\@ one iteration of ISTA executes a linear and then a non-linear operation and thus can be recast into a network layer; by stacking the layers a deep network is formed. The network is subsequently trained using paired inputs and outputs by back-propagation to optimize the parameters $\bW_e,\bW_t$ and $\lambda$. The trainable parameters in the network are colored in blue.}\label{fig:unroll_lista}
\end{figure*}

To unroll ISTA, iteration~\eqref{eq:unroll_ista_iter} can be recast into a single network
layer as depicted in Fig.~\ref{fig:unroll_lista}. This layer comprises a series of analytical operations (matrix-vector
multiplication, summation, soft-thresholding), which is of the same nature as a
neural network. A diagram representation of one iteration step reveals its
resemblance to a single network layer. Executing ISTA $L$ times can be
interpreted as cascading $L$ such layers, which essentially forms an
$L$-layer deep network. Note that, in the unrolled network an implicit
substitution of parameters has been made: $\bW_t=\bI-\frac{1}{\mu}\bW^T\bW$ and
$\bW_e=\frac{1}{\mu}\bW^T$.

After unrolling ISTA into a network, named Learned ISTA (LISTA), the network is
trained through back-propagation using real datasets to optimize the parameters
$\bW_t$, $\bW_e$ and $\lambda$. Training is performed in a supervised manner,
meaning that for every input vector $\by^t\in\mathbb{R}^n, t=1,..,T$, its
corresponding sparse output $\bx^{*t}\in\mathbb{R}^m, t=1,..,T$ is known (by choosing an approriate $\lambda$ which is fixed and might be different from the learned value). The
sparse codes $\bx^{\ast t}$ can be determined, for example, by executing ISTA
when $\bW$ is known. Inputting the vector $\by^t$ into the network results in a
predicted output $\hat{\bx}^t(\by^t;\bW_t, \bW_e,\lambda )$. The network
training loss function is formed by comparing the prediction with the known
sparse output $\bx^{*t}$:
\begin{equation}
    \ell(\bW_t,\bW_e,\lambda)=\frac{1}{T}\sum_{t=1}^{T}\left\|\hat{\bx}^t\left(\by^t;\bW_t, \bW_e,\lambda\right)- \bx^{*t}\right\|_2^2.
\end{equation}
The network is trained through loss minimization, using popular gradient-based
learning techniques such as stochastic gradient descent, to learn the unknown
parameters $\bW_t$, $\bW_e$ and $\lambda$~\cite{lecun_efficient_2012}.

The learned version, LISTA, may achieve higher efficiency compared to ISTA\@.
It is also useful when it is analytically hard to determine $\bW$. Furthermore,
it has been shown empirically that, the number of layers $L$ in (trained) LISTA
can be an order of magnitude smaller than the number of iterations required for
ISTA to achieve convergence corresponding to a new observed input, thus
dramatically boosting the sparse coding efficiency~\cite{gregor_learning_2010}.
In practice, $\bW_t$, $\bW_e$ and $\lambda$ can be untied and vary in each
layer.

\subsection{Towards Theoretical Understanding of Algorithm Unrolling}\label{ssec:unroll_theory}
While LISTA empirically achieves much higher efficiency compared to ISTA,
theoretical understanding of its behavior is still lacking. In particular,
conditions offering convergence guarantees, rates of convergence, structure of
the optimal solutions, factors contributing to higher empirical performance,
are some of the important problems that need to be answered. In recent years,
researchers have actively analyzed LISTA and its variants and made significant
progress.

Xin {\it et al.}~\cite{xin2016maximal} study the unrolled Iterative Hard
Thresholding (IHT) algorithm, which has been widely applied in various sparsity
constrained estimation problems. IHT largely resembles ISTA except that an
$\ell^0$ norm is employed instead of the $\ell^1$ norm. Formally, IHT
solves the following optimization problem instead of~\eqref{eq:unroll_l1_min}:
\begin{equation}
	\min_{\bx\in\mathbb{R}^m}\frac{1}{2}\|\by-\bW\bx\|_2^2\quad\text{subject to}\quad\|\bx\|_0\leq k,\label{eq:unroll_l0_min}
\end{equation}
for some positive integer $k$. IHT essentially performs the following iterations:
\begin{align}
	\bx^{l+1}&=\cH_k\left\{\left(\bI-\frac{1}{\mu}\bW^T\bW\right)\bx^l+\frac{1}{\mu}\bW^T\by\right\},\quad l=0,1,\dots,\label{eq:unroll_iht_orig}
\end{align}
where $\mu$ is a positive parameter controlling the step size, and $\cH_k$ is
the hard-thresholding operator which zeros out all but the largest (in
magnitude) $k$ coefficients of the vector. Similar to LISTA,
in~\cite{xin2016maximal} $\bW_t=\bI-\frac{1}{\mu}\bW^T\bW$ and
$\bW_e=\frac{1}{\mu}\bW^T$ are used instead of $\bW$ itself, which reduces the
iteration~\eqref{eq:unroll_iht_orig} into the following version:
\begin{align}
	\bx^{l+1}&=\cH_k\left\{\bW_t\bx^l+\bW_e\by\right\},\quad l=0,1,\dots.\label{eq:unroll_iht_repar}
\end{align}

The authors then prove that, a necessary condition for the
iteration~\eqref{eq:unroll_iht_repar} to recover a feasible solution
to~\eqref{eq:unroll_l0_min} is that, the weight coupling scheme $\bW_t=\bI-\bW_e\bW$
is satisfied, for some $\bW$ introduced here as a free variable. Note that this formula is reminiscent of the implicit re-parameterization scheme: $\bW_t=\frac{1}{\mu}\bW^T\bW$. Furthermore, under the weight coupling constraint, when
$\bW$ and $\bW_e$ additionally obey technical conditions, the
iteration~\eqref{eq:unroll_iht_repar} recovers the underlying solution at a
linear rate. In particular, $\bW$ needs to satifsy certain Restricted-Isometry-Property (RIP) conditions. The authors also articulate the exact forms of $\bW_e$ in order to ensure linear convergence. Compared to classical IHT, the
iteration~\eqref{eq:unroll_iht_repar} necessitates a much more relaxed
requirement on the RIP condition, which implies that the unrolled network is
capable of recovering sparse signals from dictionaries with relatively more
coherent columns.

Chen {\it et al.}~\cite{chen_theoretical_2018} study the LISTA network with
layer specific parameters ${\{\bW_t^l,\bW_e^l,\lambda^l\}}_{l\in\mathbb{N}}$,
and prove that, if LISTA recovers the underlying sparse solution and if the
sequence $\bW^1_t,\bW^2_t,\dots$ is bounded, then a similar weight
coupling scheme must be satisfied asymptotically:
\begin{align*}
	\bW_t^l-(\bI-\bW_e^l\bW)&\rightarrow 0,\quad\text{as}\quad l\rightarrow\infty,\\
	\lambda^l&\rightarrow 0,\quad\text{as}\quad l\rightarrow\infty.
\end{align*}
When LISTA adopts the weight coupling parametrization $\bW=\bI-\bW_e^l\bW$, and
employs a dedicated support selection technique, the resulting network is
called LISTA-CPSS, where ``CP'' stands for weight coupling and ``SS'' stands
for support selection. Chen {\it et al.} prove that, if the underlying solution
$\bx^\ast$ is sufficiently sparse and bounded, then the sequence
${\{\bW_e^l,\lambda^l\}}_{l\in\mathbb{N}}$ can be chosen according to certain values such that LISTA-CPSS recovers
$\bx^\ast$ at a linear rate. As a follow-up, Liu {\it et
al.}~\cite{liu2018alista} introduce certain mutual coherence conditions and
analytically characterize optimal network parameters based on those conditions.
Similar to the networks with trained weights, networks adopting analytic
weights converge at a linear rate, which implies that analytic weights can be
as efficient as learned weights. In addition, analytic weights are of much
lower dimensionality compared to trained weights. However, determining the
analytic weights can be a nontrivial task as typically another optimization
problem has to be solved.

\subsection{Unrolling Generic Iterative Algorithms}\label{ssec:unroll_general}
\begin{figure*}
	\centering
	\includegraphics[width=\textwidth]{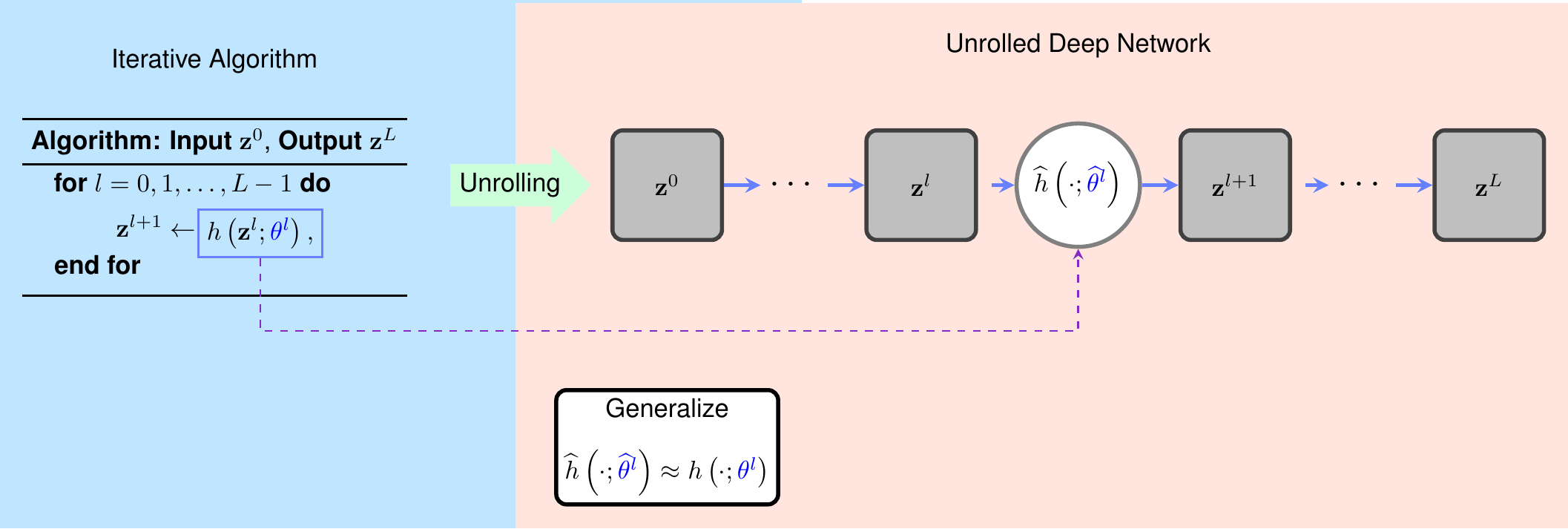}
	\caption{Illustration of the general idea of algorithm unrolling: given an iterative algorithm, we map one iteration (described as the function $h$ parametrized by $\theta^l,l=0,\dots,L-1$) into a single network layer, and stack a finite number of layers to form a deep network. Feeding the data forward through an $L$-layer network is equivalent to executing the iteration $L$ times (finite truncation). The parameters $\theta^l,l=0,\dots,L-1$ are learned from real datasets by training the network end-to-end to optimize the performance. They can either be shared across different layers or varying from layer to layer. The trainable parameters are colored in blue.}\label{fig:unroll_general}
\end{figure*}

Although the initial motivation of Gregor {\it et al.}'s
work~\cite{gregor_learning_2010} was to increase the efficiency of sparse
coding techniques via training, the underlying principles could be easily
generalized. More specifically, provided with a certain iterative algorithm, we
can unroll it into a corresponding deep network, following the procedures
depicted in Fig.~\ref{fig:unroll_general}. The first step is to identify the
analytic operations per iteration, which we represent abstractly as an $h$
function, and the associated parameters, which we denote collectively as
$\theta^l$. The next task is to generalize the functional form of $h$ into a
more generic version $\widehat{h}$, and correspondingly expand the parameters $\theta^l$
into an enlarged version $\widehat{\theta}^l$ if necessary. For instance,
in LISTA the parameter $\bW$ is substituted with $\bW_t$ and $\bW_e$ through
the formula $\bW_t=\bI-\frac{1}{\mu}\bW^T\bW$ and $\bW_e=\frac{1}{\mu}\bW^T$.
After this procedure each iteration can be recast into a network layer in the
same spirit as LISTA\@. By stacking the mapped layers together we obtain a deep
network with undetermined parameters, and then obtain optimal parameters
through end-to-end training using real world datasets.

The exact approach to generalize $h$ and $\theta^l$s towards $\widehat{h}$ and
$\widehat{\theta}^l$s is largely case specific. An
extreme scenario is to strictly follow the original functional forms and
parameters, i.e., to take $\widehat{h}=h$ and
$\widehat{\theta}^l=\theta^l,\forall l$. In this way, the trained network
corresponds exactly to the original algorithm with finite truncation and
optimal parameters. In addition to efficiency enhancement thanks to
training~\cite{gregor_learning_2010}, the unrolled networks can aid with
estimating structured parameters such as filters~\cite{solomon2019sparcom} or
dictionaries~\cite{wang_deep_2015} which are hard to design either analytically
or by handcrafting. Alternatively, some operations may be
replaced with a stand alone deep neural network such as Convolutional
Neural Network (CNN) or Recurrent Neural Network (RNN). For instance,
in~\cite{hauptmann2018model} the authors replace a proximal gradient
update step with a CNN\@. In addition, the parameters can be layer
specific instead of being shared across different layers. For instance,
in~\cite{adler2018learned} the authors plug in a CNN in each iteration
step (layer) and allow the network parameters to differ. As it is, networks
with shared parameters generally resemble RNN, while those with layer specific
parameters mimic CNN, especially when there are convolutional structures
embedded in layer-wise operations. It is important to note that such custom modifications may potentially
sacrifice certain conceptual merits such as invalidating convergence guarantees, introducing departures from the original iterative algorithms, or undermining the interpretability to some extent; nevertheless, they are practically beneficial and
critical for performance improvement because the representation capacity of the
network can be significantly extended.

In addition to performance and efficiency benefits, unrolled networks can
potentially reduce the number of parameters and hence storage footprints.
Conventional generic neural networks typically reuse essentially the same
architectures across different domains and thus require a large amount of
parameters to ensure their representation power. In contrast, unrolled networks
generally deliver significantly fewer parameters, as they implicitly transfer
problem structures (domain knowledge) from iterative algorithms to unrolled
networks and their structures are more specifically tailored towards target
applications. These benefits not only ensure higher efficiency, but also
provide better generalizability especially under limited training
schemes~\cite{yang2018admm}, as will be demonstrated through concrete
examples in Section~\ref{sec:unroll_application}.

The unrolled network can share inter-layer parameters, thus resembling a
    RNN, or carry over layer-specific parameters. Although parameter sharing
    could further reduce parameters, the RNN-like architecture significantly
    increases the difficulty in training. In particular, it may suffer from
    gradient explosion and vanishing problems. On the other hand, using layer
    specific parameters may lead to deviation from the original
    iterative algorithm and may no longer inherit the convergence guarantees.
However, the networks can have larger capacity and become easier to train.

Another importatn concern when designing unrolled networks is to determine
    the optimal number of layers, which corresponds to the number of iterations
    in the iterative algorithm. Most exisiting approaches generally treat it as
    a hyperparameter and determine it through cross-validation. However, recent theoretical breakthroughts (some of them reviewed in
    Section~\ref{ssec:unroll_theory}) regarding convergence rate of the unrolled network can shed some
light and guide future efforts.

\section{Deep Algorithm Unrolling for Biomedical Imaging}\label{sec:unroll_application}
In recent years, algorithm unrolling has found wide applications in many areas
of biomedical imaging. Traditionally, iterative algorithms have played a
dominant role in solving various biomedical imaging problems. Recently,
learning-based approaches, especially deep neural networks, have become
increasingly popular. Algorithm unrolling bridges between well-grounded
iterative algorithms and contemporary deep networks and has attracted growing
research attention. In this section, we discuss how algorithm unrolling
can be successfully applied towards biomedical image synthesis via several
concrete cases.

\subsection{Applications of Unrolling in Computed Tomography}\label{ssec:unroll_tomography}
In general, Computed Tomography (CT) refers to a class of imaging techniques
that reconstruct the original signal out of its directional projections
(slices). For instance, in X-ray CT, a mobilized X-ray source rotates around
the patient to emit narrow beams of X-rays, which are then received by X-ray
detectors located opposite of the X-ray source. The X-rays pass through the patient to create image slices. In it simplest form, this procedure can be described by the well-known Radon transform. In 2D, the Radon transform is given by the following formula:
\begin{equation*}
	p(\xi,\phi)=\int f(x, y)\delta(x\cos(\phi)+y\sin(\phi)-\xi)\mathrm{d}x\mathrm{d}y,
\end{equation*}
where $f$ is the imaged signal and $\delta$ is the Dirac delta function. The
function $p$ is often referred to as a \emph{sinogram}. To invert the Radon
transform, the classical central slice theorem plays a critical role as it
relates the Fourier transform of the original signal and its sinogram. This
relationship gives rise to the Filtered Back Projection (FBP) algorithm: the
sinogram is first filtered by a so-called ramp filter, and then back-projected
to obtain the original signal. Practical challenges of CT include short
scanning time, low-dose, motion and noise.

In an early work that employs deep CNN for CT~\cite{jin_deep_2017}, the authors observe the architectural similarity between the
popular U-net~\cite{Ronneberger_unet_2015} and the unrolled ISTA network. They then
combine FBP with U-net to construct a deep network called FBPConvNet. Through
extensive experimental studies on sparse-view X-ray CT reconstruction,
FBPConvNet offers clear benefits over traditional iterative reconstruction
techniques such as the Total-Variation (TV) regularization approach. In
particular, FBPConvNet achieves over 3dB Signal-to-Noise Ratio (SNR) improvement over TV on a
biomedical dataset that comprises 500 real in-vivo CT images, and reduces the running time from several minutes to less
than a second. Compared with conventional deep networks such as residual
networks, FBPConvNet also achieves more than 1dB higher SNR on the same dataset.\@.

\begin{figure*}
	\includegraphics[width=\textwidth]{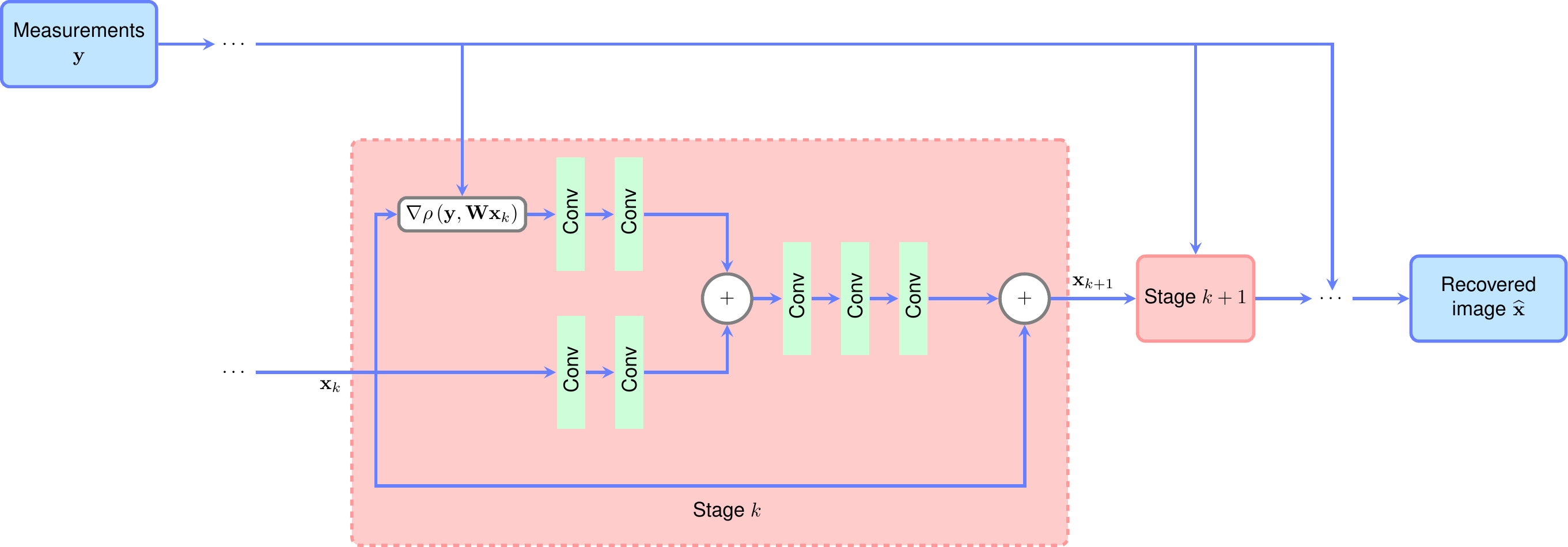}
	\caption{Representation of DGD~\protect\cite{hauptmann2018model}: each stage comprises a gradient descent iteration with the proximal operator replaced by a CNN\@. The CNN takes as input a concatenation of $\bx_k$ and the gradient $\nabla\rho\left(\by,\bW\bx_k\right)$. It also adopts a skip connection from $\bx_k$ to $\bx_{k+1}$, which implies it learns a residual mapping.}\label{fig:unroll_dgd}
\end{figure*}

An emerging CT technique is Photo Acoustic Tomography (PAT) that provides high
resolution 3D images by sensing laser-generated ultrasound.
In~\cite{hauptmann2018model} the authors develop a deep learning
technique to reconstruct high-resolution 3D images from restricted
photo acoustic measurements. Modeling the acoustic signal via the wave equation,
the initial pressure $\bx$ and the measured photo acoustic signal $\by$
satisfy a linear mapping
\[
	\by=\bW\bx,
\]
and thus PAT can be carried out by solving a linear inverse problem. A typical
approach is to solve the following regularized problem:
\begin{equation}
	\min_{\bx}\rho(\by,\bW\bx)+\lambda\cR(\bx),\label{eq:unroll_pat}
\end{equation}
where $\rho(\cdot,\cdot)$ is a metric measuring the data consistency, $\cR$ is
a regularizer and $\lambda$ is the regularization coefficient. A simple
approach to solving~\eqref{eq:unroll_pat} is the Proximal Gradient Descent
(PGD) algorithm, which performs the following iterations:
\begin{equation}
	\bx_{k+1}\gets\mathrm{prox}_{\cR}\left(\bx_k-\gamma_{k+1}\nabla\rho(\by,\bW\bx_k);\lambda\gamma_{k+1}\right),\label{eq:unroll_pgd}
\end{equation}
where $\gamma_{k+1}>0$ controls the step size. Hauptmann {\it et al.} replace
the proximal operator with a CNN and learn the arithmetic operations
(subtraction, multiplication) rather than fixing it according to the gradient
descent rule~\eqref{eq:unroll_pgd}. In other words, they propose to unroll the
following update procedures:
\[
	\bx_{k+1}=G_{\theta_k}\left(\nabla\rho(\by,\bW\bx_k),\bx_k\right),
\]
where $G$ denotes a CNN and $\theta_k$ are its parameters. The corresponding
unrolled network is depicted visually in Fig.~\ref{fig:unroll_dgd}. The network
is dubbed Deep Gradient Descent (DGD). To train the network, a stage-wise
scheme is employed: at the $k$-th stage, $\theta_k$ is optimized by minimizing the
Mean-Square-Error (MSE) loss between $\bx_k$ and groundtruth $\bx^\ast$ while
holding other $\theta_j (j\neq k)$'s fixed. Experimentally, DGD achieves nearly
1dB improvement in Peak Signal-to-Noise Ratio (PSNR) over U-Net and more than
3dB over TV reconstruction on an in-vivo dataset of a human
palm. Furthermore, compared with conventional deep networks such as U-Net, DGD proves its superior robustness
against perturbations of measurement procedures or targets, by showing only
slight deterioration in reconstruction quality under small perturbations such
as varying sub-sampling patterns, noise levels, or deviations in sound speed.

\begin{figure*}
	\centering
	\includegraphics[width=\textwidth]{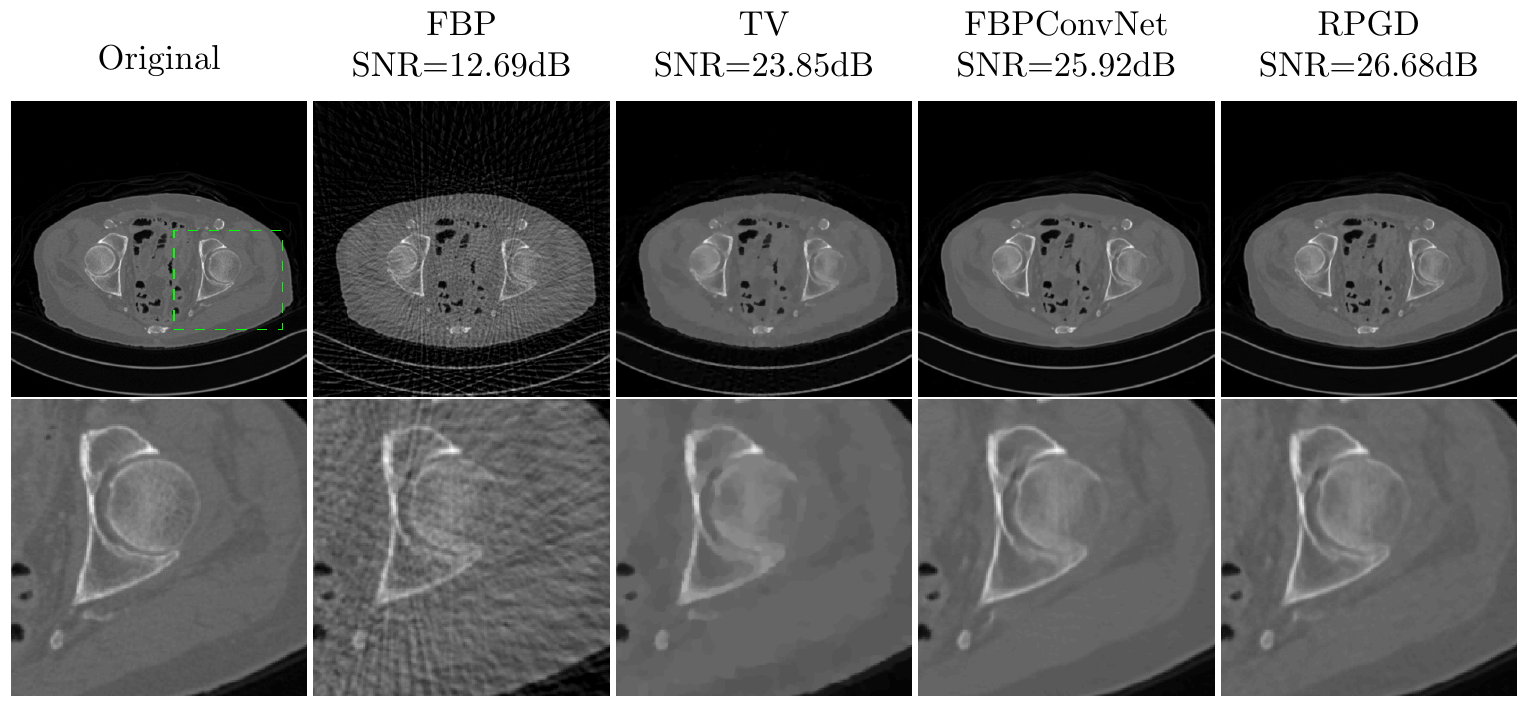}
	\caption{Experimental comparisons between RPGD~\protect\cite{gupta2018cnn} and state-of-the art reconstruction algorithms (including FBPConvNet from~\protect\cite{jin_deep_2017}) over a test sample from noiseless measurements with 45 views (x16 dosage reduction). The SNR values are also included for quantitative comparisons. The first row shows the full images and the second row shows their magnified portions, respectively. Images are courtesy of Michael Unser at EPFL~\protect\cite{gupta2018cnn}.}\label{fig:unroll_rpgd}
\end{figure*}

In a closely related technique~\cite{gupta2018cnn}, Gupta {\it et al.}
unroll the PGD algorithm, and similarly replace the proximal operator with
a CNN.\@ However, they do not learn the arithmetic operations and instead
follow the update rule in~\eqref{eq:unroll_pgd}. In addition, they introduce a
relaxation to the PGD update, in order to maintain its convergence guarantee
even when the proximal operator is substituted with a CNN\@. The unrolled
network, dubbed Relaxed Projected Gradient Descent (RPGD), yields substantially
improved CT recovery, as depicted in Fig.~\ref{fig:unroll_rpgd}.

Low-dose X-ray CT is an intriguing technique as X-ray CT causes potential
cancer risks due to radiation exposure. A drawback of this approach, however,
is the low SNR of the projections due to noise, and thus power imaging
techniques are required to retrieve high-quality reconstructed images. To
overcome the challenge of noisy measurements, Kang {\it et
al.}~\cite{kang2018deep} build a Wavelet Residual Network (WavResNet) based on
a theory of deep convolutional framelet for deep learning-based denoising.
Suppose we seek to estimate a ground-truth signal
$\bx^\ast\in\mathbb{R}^n$ from its noisy measurement $\by\in\mathbb{R}^n$:
\[
	\by=\bx^\ast+\bn,
\]
where $\bn\in\mathbb{R}^n$. A traditional approach is frame-based
denoising, which shrinks the frame coefficient $\balpha$ of the signal by solving:
\[
	\min_{\bx,\balpha}\frac{\mu}{2}\|\by-\bx\|_2^2+\frac{1-\mu}{2}\left\{\|\bW\bx-\balpha\|_2^2+\lambda\|\balpha\|_1\right\},
\]
where $\lambda,\mu>0$ are the regularization coefficients and $\bW$ is the
analysis operator of the frame. The corresponding proximal update
equation is given by:
\begin{equation}
	\bx_{k+1}=\mu\by+(1-\mu)\bW^T\cS_\lambda\left(\bW\bx_k\right),\label{eq:unroll_frame_proximal}
\end{equation}
where $\cS_\lambda$ is the soft-thresholding operator with thresholding value
$\lambda$~\cite{li2014wavelet,dong2017image}.

Kang {\it et al.}~\cite{kang2018deep} claim that the term
$\bW^T\cS_\lambda(\bW\bx_k)$ can be regarded as a CNN based on the deep
convolutional framelet theory~\cite{ye2018deep}. In short, the analysis and
synthesis operators of a frame can be concatenated to form an encoder-decoder
layer structure which corresponds to a CNN, and the shrinkage operation can be
implicitly obtained by reducing the channels in the CNN\@. The update
rule~\eqref{eq:unroll_frame_proximal} is then substituted with
\begin{equation}
	\bx_{k+1}=\mu\by+(1-\mu)\cQ\left(\bx_k;\phi,\widetilde{\phi}\right)\label{eq:unroll_framelet}
\end{equation}
where $\cQ$ is the CNN corresponding to the framelet, with undetermined filter
coefficients $\phi,\widetilde{\phi}$ corresponding to the primal and dual
frames, respectively. The unrolled networks are formed out of the update
rules~\eqref{eq:unroll_framelet} and the unknown parameters are learned; the
parameters are shared across different iterations giving rise to a RNN-like
architecture. In addition, Kang {\it et al.}~\cite{kang2018deep} slightly
modify the update rules~\eqref{eq:unroll_framelet} to ensure convergence.
Through extensive experimental results, WavResNet shows its advantage
at noise reduction and preserving texture details of the organ while
maintaining the lesion information over state-of-the-art techniques.

A related technique that also focuses on low-dose CT reconstruction is the work
by Adler {\it et al.}~\cite{adler2018learned}, where the Proximal-Dual Hybrid
Gradient (PDHG) algorithm is generalized and unrolled into a learnable network.
Technically, the PDHG algorithm solves problems of the following form:
\begin{equation}
	\min_{\bx}\cP(\cK(\bx))+\cD(f),\label{eq:unroll_primal_dual}
\end{equation}
where $\cK$ is an operator over signal $\bx$ and $\cP, \cD$ are functionals on
the primal and dual spaces, respectively.
Problem~\eqref{eq:unroll_primal_dual} has many interesting instances; in
particular, the TV-regularized CT can be instantiated by letting
$\cK(\bx):=[\cR(\bx),\nabla\bx]$,
$\cP([\bh_1,\bh_2]):=\|\bh_1-\by\|_2^2+\|\bh_2\|_1$ and $\cD:=0$, which gives
rise to the following optimization problem:
\[
	\min_{\bx}\left\|\cR(\bx)-\by\right\|_2^2+\lambda\|\nabla\bx\|_1,
\]
where $\cR$ is the Radon transform operator, $\nabla$ is the gradient operator,
and $\lambda>0$ is a regularization coefficient. The PDHG algorithm essentially performs the following iterations for $k=1,2,\dots$:
\begin{align*}
	\bz_{k+1}&\gets\mathrm{prox}_{\cP^\ast}\left(\bz_k+\sigma\cK(\bar{\bx}_k);\sigma\right),\\
	\bx_{k+1}&\gets\mathrm{prox}_{\cD}\left(\bx_k-\tau{[\partial\cK(\bx_k)]}^\ast(\bz_{k+1});\tau\right),\\
	\bar{\bx}_{k+1}&\gets\bx_{k+1}+\gamma(\bx_{k+1}-\bx_k),
\end{align*}
where $\ast$ is the Fenchel conjugate and $\partial$ is the Fr\'{e}chet
derivative. For definitions of these terms, see~\cite{rockafellar1970convex}.
The operator $\mathrm{prox}_\cF(\cdot;\lambda)$ is the proximal operator over
functional $\cF$ and parameter $\lambda$, given by
\[
	\mathrm{prox}_{\cF}(\bx;\lambda)=\arg\min_{\bx'}\cF(\bx')+\frac{1}{2\lambda}\|\bx'-\bx\|_2^2.
\]

In a similar spirit to~\cite{hauptmann2018model}, Adler {\it et al.} generalize
the update rules by replacing the proximal operators with CNNs, and
substituting the arithmetic operations (summation, multiplication, etc) as
learnable operations, giving rise to the following iterations:
\begin{align*}
	\bz_k&\gets\Gamma_{\theta_k^d}\left(\bz_{k-1},\cK\left(\bx_{k-1}^{(2)}\right),\by\right),\\
	\bx_k&\gets\Lambda_{\theta_k^p}\left(\bx_{k-1},{\left[\partial\cK\left(\bx_{k-1}^{(1)}\right)\right]}^\ast\left(\bz_k^{(1)}\right)\right),
\end{align*}
where $\Lambda_{\theta_k^p}$ and $\Gamma_{\theta_k^d}$ are CNNs replacing
primal and dual operators, with parameters $\theta_k^p$ and $\theta_k^d$,
respectively. The notation $\bx^{(1)}$ and $\bx^{(2)}$ selects first and second
blocks of coefficients in $\bx$ to permit ``memory'' across iterations. A deep network,
dubbed Learned Primal-Dual, can then be formed by stacking iterations (layers)
together. The parameters $\theta_k^p$ and $\theta_k^d$ are learned by
minimizing the MSE loss.

Compared with FBP, TV reconstruction and conventional deep networks such as
U-Net, Learned Primal-Dual obtains more than 6dB improved PSNR for the
Shepp-Logan phantom, and over 2dB improvement on human phantoms on average. In terms of
Structural Similarity Index (SSIM), it also outperforms its competitors by a
large margin. The runtime is on the same order as FBP, rendering it amenable to
time-critical applications.

Another technique for low-dose CT reconstruction is the BCD-Net~\cite{chun2018deep,chun2019bcd}, where the authors generalize the block coordinate descent algorithm for model-based inversion, by integrating a CNN as a learanble denoiser. As a follow-up of the learned primal-dual network, Wu~\cite{wu_end--end_2018}
concatenates it with a detection network, and applies joint fine-tuning after
training both networks individually. Their jointly fine-tuned network
achieves comparable performance with the detector trained on the fully-sampled
data and outperforms detectors trained on the reconstructed images.

\subsection{Unrolling in Super-resolution Microscopy}\label{ssec:unroll_microscopy}
Optical microscopy generates magnified images of small objects by using visible
light and a system of lenses, thus enabling different research fields like
cell biology and microbiology. However, the spatial resolution is limited by
the physics of light, which thereby poses a hard limit on resolution. The
diffraction of light makes structures too blurry to be resolved once they are
smaller than approximately half the wavelength of
light~\cite{huang2010super,solomon2019sparcom}. This limitation is circumvented
using photo-activated fluorescent molecules. In 2-D Single Molecule
Localization Microscopy (SMLM), a sequence of diffraction-limited images,
produced by a sparse set of emitting fluorophores with minimally overlapping
Point Spread Functions (PSFs) is acquired, allowing the emitters to be
localized with high  precision~\cite{dardikman-yoffe_learned_2020}. However,
the need for low emitter-density results in poor temporal resolution.

In~\cite{dardikman-yoffe_learned_2020}, the authors unroll the SPARsity based
super-resolution COrrelation Microscopy (SPARCOM)~\cite{solomon2019sparcom}
method to incorporate domain knowledge, resulting in Learned SPARCOM
(LSPARCOM). Specifically, LSPARCOM aims to recover a single $M \times M$ high
resolution image $\bX$, corresponding to the locations of the emitters on a
fine grid, from a set of $T$ low-resolution frames of size $N \times N$ with
$N<M$. A single frame taken at time $t$ is denoted as
$\mathbf{Y}(t)$. The method exploits the sparse nature of the fluorophores
distribution, alongside a statistical prior of uncorrelated emissions.
The sparsity assumption implies that the overall number of
emitter-containing-pixels is significantly smaller than the total number of
pixels in the high resolution grid, yet, as opposed to classical methods for
SMLM, it accounts for scenarios with overlapping PSFs formed by adjacent
emitters.

The authors formulate a sparse recovery problem on the temporal variance image given by:
\begin{equation}
    \mathbf{g}_Y = \bW\bx, \label{eq:unroll_LSPARCOM}
\end{equation}
where $\mathbf{g}_Y\in\mathbb{R}^{N^2}$ comprises temporal variances of the set
of $T$ low-resolution frames, $\bW$ is a dictionary matrix based on the
PSF, and $\bx$ comprises variances of the emitter fluctuation
on a high-resolution grid. The unknown $\bx$ can be determined by solving a
sparsity constrained optimization problem similar to~\eqref{eq:unroll_l1_min}
with an additional non-negativity constraint over $\bx$, which in turn can be
solved via the ISTA algorithm presented in Section~\ref{ssec:unroll_ista}. The
algorithm can be further unrolled into a network whose architecture resembles
LISTA\@. Different from LISTA, the authors propose the following
proximal operator instead of the soft-thresholding operator defined
in~\eqref{eq:unroll_soft_thresh_op}:
\begin{equation}
	\cS_{\alpha,\beta}^+(x) \overset{\triangle}{=}\frac{\max\{0,x\}}{1+\exp{\left[- \beta(|x|-\alpha)\right]}},\label{eq:unroll_hard_thresh_op}
\end{equation}
where $\alpha$ and $\beta$ are hyper-parameters. When applied to vectors and
matrices, $\cS_{\alpha,\beta}^+(\cdot)$ operates element-wise. This is a 
sigmoid-based approximation of the positive hard thresholding operator, which
is the proximal operator for the $\ell^0$ norm, that accounts for increased
sparsity. Furthermore, it performs one-sided thresholding due to the
non-negativity constraint.

For training, the authors introduce a compatible loss function given by:
\begin{equation}
	Loss=\frac{1}{M^2}\sum_{i,j=1}^{M}\bB(i,j)|\hat{\bX}(i,j)-\bX^{*}(i,j)|^2+\lambda\left[1-\bB(i,j)\right]|\hat{\bX}(i,j)|,
\end{equation}
where $\hat{\bX}$ is the output of the network, $\bX^{*}$ is the ground truth
image, and $\bB$ is a binary mask, created by binarizing $\bX^{*}$, which
eliminates entries that do not contain emitters. This encourages the network to
output signals coming only from emitters while suppressing the background.

\begin{figure}
	\centering
	\includegraphics[width=0.8\textwidth]{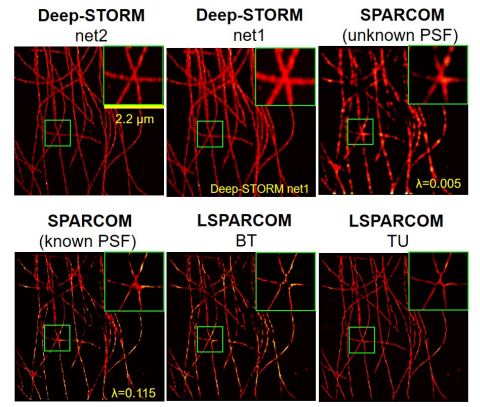}
	\caption{Performance evaluation for experimental tubulin sequence, composed of 500 high-density frames [35]. A difficult area for reconstruction is shown magnified in the green box. From top left to bottom right: reconstructions using Deep-STORM architecture trained on different datasets (net1 and net 2), SPARCOM reconstruction with unknown PSF, SPARCOM reconstruction using explicit knowledge of the PSF, reconstructions using LSPARCOM architecture trained on different datasets (BT, TU). Figure reproduced from~\protect\cite{dardikman-yoffe_learned_2020} with authors' permission.}\label{fig:unroll_res_lsparcom}
\end{figure}

Experimental results across different test sets show that the unrolled network
can obtain super-resolution images from a small number of high emitter density
frames without knowledge of the optical system. All the methods that are
compared were evaluated on the quality of the reconstruction based on the exact
same input frames, while the input for LSPARCOM is a single image, which is
constructed by calculating the temporal variance of all the high density
frames. Fig.~\ref{fig:unroll_res_lsparcom} provides en example for performance evaluation of SPARCOM, LSPARCOM, and Deep-STORM~\cite{nehme2018deep}. Deep-STORM is another learning based method, devised for the problem at hand. Two examples are brought for each method, one that incorporates prior knowledge  (Deep-STORM net2 and LSPARCOM TU which were trained for similar imaging parameters as in the test set and SPARCOM with known PSF), and the second without prior knowledge. While all three methods achieve good results when incorporating prior knowledge, LSPARCOM is superior when generalizing to new data.
Quantitatively speaking, SPARCOM reconstruction took 39.32 sec, LSPARCOM reconstruction took 10.8 sec, and Deep-STORM took 280.38 sec for 500 128 × 128 input frames. Moreover, LSPARCOM is highly parameter efficient: 9058 parameters
in 10 folds of LSPARCOM vs. 1.3M trainable parameters in Deep-STORM.\@ Consequently, the network achieves runtime savings compared to other learning-based methods.

\subsection{Applications of Unrolling in Ultrasound}\label{ssec:unroll_ultrasound}
An important ultrasound-based modality is Contrast Enhanced UltraSound\\
(CEUS)~\cite{furlow2009contrast}, which allows the detection and visualization
of small blood vessels. The main idea behind CEUS is the use of encapsulated
gas microbubbles, serving as Ultrasound Contrast Agents (UCAs). UCAs  are
injected intravenously and can flow throughout the vascular system due to their
small size. To visualize them, strong clutter signals originating from
stationary or slowly moving tissues must be removed, as they introduce
significant artifacts in the resulting images. The latter poses a major
challenge in ultrasonic vascular imaging~\cite{van2019deep}.

In~\cite{solomon_deep_2018}, the authors suggest a method for clutter
suppression by applying the well-known low-rank and sparse matrix decomposition
technique, a.k.a Robust Principal Component Analysis (RPCA), and unrolling the
corresponding algorithm into a deep network named Convolutional rObust
pRincipal cOmpoNent Analysis (CORONA). Specifically, they model the acquired
contrast enhanced ultrasound signal as a combination of a low-rank matrix
(tissue) and a sparse outlier signal (UCAs).

In UltraSound (US) imaging, a series of pulses are transmitted to the imaged medium. The resulting echoes from the medium are received in each transducer element and then combined in a process called beamforming to produce a focused image. If we denote the vertical and axial
coordinates as $x$, $z$, and the frame number as $t$, the observed signal from
a specific point in time-space, denoted as $Y(x,z,t)$, can be described as the
sum of echoes returned from the blood and CEUS signal $S(x,z,t)$ as well as
from the tissue $L(x,z,t)$, contaminated by additive noise $N(x,z,t)$. When
acquiring a series of movie frames $t = 1,\dots, T$, and stacking them as
columns in a matrix $\mathbf{Y}$, the relation between the measurements and the
signal scan be described via:
\begin{equation}
       \mathbf{Y}=\bH_1\bL+\bH_2\bS+\bN,\label{eq:unroll_RPCA}
\end{equation}  
where $\bH_1$ and $\bH_2$ are measurement matrices of appropriate dimensions.
Similar to $\mathbf{Y}$, $\bL$ and $\bS$ are matrices whose columns contain
per-frame data. The tissue matrix $\bL$ can be described as a low-rank matrix,
due to its high spatio-temporal coherence. The CEUS echoes matrix $\bS$ is
assumed to be sparse, as blood vessels typically sparsely populate the imaged
medium. Under the assumptions on the matrices $\mathbf{L}$ and $\mathbf{S}$,
their recovery can be obtained by solving the following convex optimization
problem:
\begin{equation}
    \min_{\bL,\bS}\frac{1}{2}\|\mathbf{Y}-(\bH_1\bL+\bH_2\bS)\|_F^2+\lambda_1\|\bL\|_*+\lambda_2\|\bS\|_{1,2},\label{eq:unroll_min_RPCA}
\end{equation}
where $\|\cdot\|_*$ stands for the nuclear norm (i.e., the sum of the singular
values of $\bL$), $\|\cdot\|_{1,2}$ stands for the mixed $\ell^{1,2}$ norm
(i.e., the sum of the $\ell^2$ norms of each row of $\bS$), and $\lambda_1$ and
$\lambda_2$ are regularization coefficients promoting low rank structure of
$\bL$ and sparsity of $\bS$, respectively.

Problem~\eqref{eq:unroll_min_RPCA} can be
solved using a generalized version of ISTA over matrices, by utilizing
the proximal mapping corresponding to the nuclear norm and mixed $\ell_{1,2}$
norm~\cite{monga2019algorithm}. In each iteration $l$, the estimations for both
$\bS$ and $\bL$ are updated as follows:
\begin{equation}
\begin{split}
    \bL^{l+1}=\cT_{\frac{\lambda_1}{\mu}}\left\{\left(\bI-\frac{1}{\mu}\bH_1^H\bH_1\right)\bL^{l}-\bH_1^H\bH_2\bS^{l}+\bH_1^H\mathbf{Y}\right\}, \\
    \bS^{l+1}=\cS_{\frac{\lambda_2}{\mu}}^{1,2}\left\{\left(\bI-\frac{1}{\mu}\bH_2^H\bH_2\right)\bS^{l}-\bH_2^H\bH_1\bL^{l}+\bH_2^H\mathbf{Y}\right\}.
\end{split}
\end{equation}
Here $\cT_{\lambda}\{\bX\}$ is the singular value thresholding operator that performs soft thresholding over the singular values of $\bX$ with threshold $\lambda$, $\cS_\lambda^{1,2}$ performs row-wise soft thresholding with parameter $\lambda$, and $\mu$ is the step size parameter for ISTA.\@

\begin{figure}
	\centering
	\includegraphics[width=\textwidth]{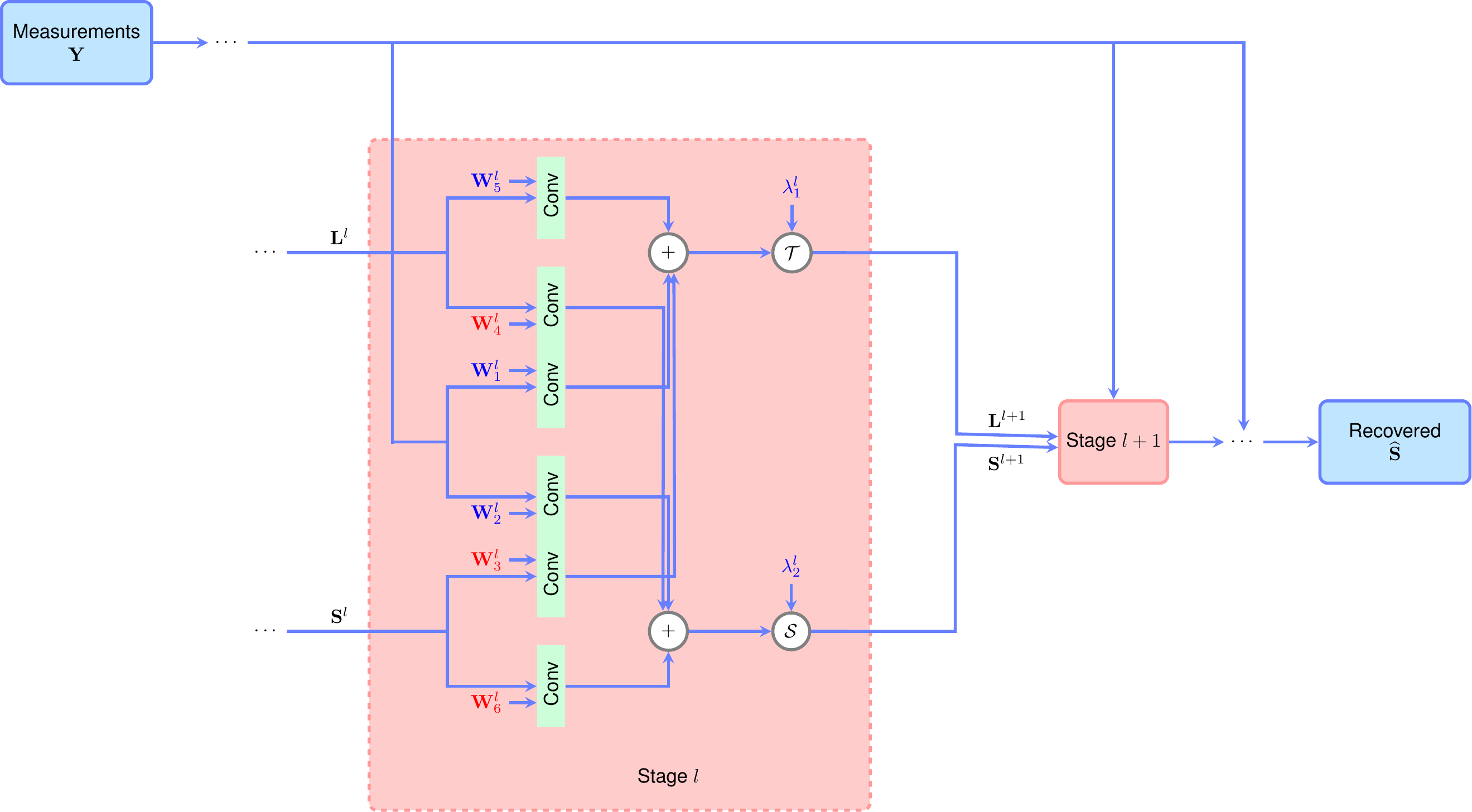}
	\caption{Diagram representation of CORONA.\@ Trainable parameters are colored in blue. The learned network draws its architecture from the iterative algorithm. Here $\mathbf{Y}$ is the input measurement matrix, and $\bS^{l}$ and $\bL^{l}$ are the estimated sparse and low-rank matrices in each layer, respectively.}\label{fig:unroll_arch_corona}
\end{figure}

To construct the unrolled network, Solomon {\it et al.} replace matrix
multiplications with convolutional layers that are subsequently fed into the
non-linear proximal mappings. The $l$-th iteration then becomes:
\begin{equation}
\begin{split}
    \bL^{l+1}=\cT_{\lambda_1^l}\left\{\bW_5^l\ast\bL^{l}+\bW_3^l\ast\bS^{l}+\bW_1^l\ast\mathbf{Y}\right\}, \\
    \bS^{l+1}=\cS_{\lambda_2^l}^{1,2}\left\{\bW_6^l\ast\bS^{l}+\bW_4^l\ast\bL^{l}+\bW_2^l\ast\mathbf{Y}\right\},
\end{split}
\end{equation}
where ${\bW_i^{l}, i=1,\dots,6}$ are a series of convolution filters that are
learned from the data in the $l$-th layer, $\ast$ acts on both spatial and temporal dimensions and performs multi-channel convolution, and
$\lambda_1^l, \lambda_2^l$ are thresholding parameters for the $l$-th layer.
The architecture is illustrated in Fig.~\ref{fig:unroll_arch_corona}, in which
$\bL^{0}$ and $\bS^{0}$ are set to some initial guess, and the output after $L$
iterations, denoted $\bS^{L}$, is used as the estimate of the desired sparse
matrix $\bS$.

The network is trained using back-propagation in a supervised manner. Training
examples consists of observed matrices along with their corresponding ground
truth $\bL^\ast$ and $\bS^\ast$ matrices (refer to~\cite{solomon_deep_2018} for more
details on how the dataset is generated). The loss function is chosen as the sum
of the MSE between $\bL^\ast$ and $\bS^\ast$ to the values predicted by the network.

\begin{figure*}
	\centering
	\includegraphics[width=\textwidth]{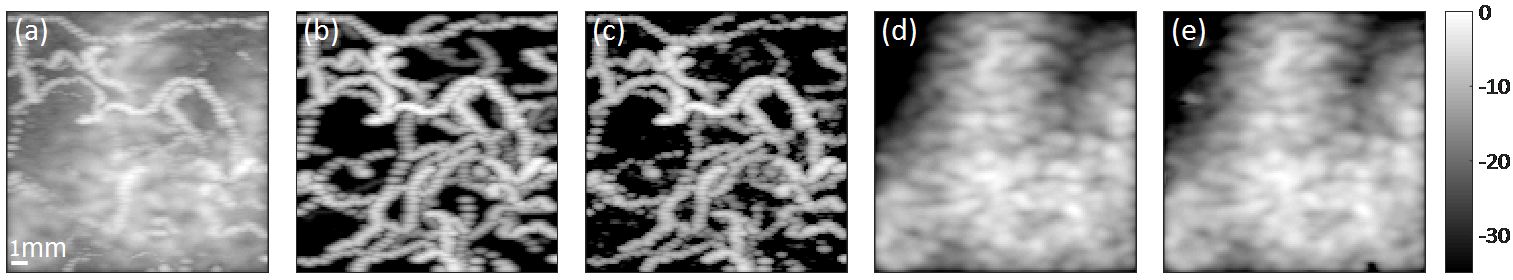}
	\caption{Sample experimental results demonstrating recovery of Ultrasound Contrast Agents (UCAs) from cluttered Maximum Intensity Projection (MIP) images. (a) MIP image of the input movie, composed from 50 frames of simulated UCAs cluttered by tissue. (b) Ground-truth UCA MIP image. (c) Recovered UCA MIP image via CORONA.\@ (d) Ground-truth tissue MIP image. (e) Recovered tissue MIP image via CORONA.\@ Color bar is in dB. Figure reproduced from~\protect\cite{solomon_deep_2018} with authors' permission.}\label{fig:unroll_corona_results}
\end{figure*}

Compared with state-of-the-art approaches, CORONA demonstrates vastly improved
reconstruction quality and has much fewer parameters.
Fig.~\ref{fig:unroll_corona_results} provides visual results, demonstrating the
power of CORONA to properly separate the low rank matrix and the sparse matrix.
Fig.~\ref{fig:unroll_corona_results}a shows the temporal maximum intensity
projection (MIP) image of the input movie (50 frames). It can be seen that the UCA
signal, depicted as randomly twisting lines, is masked considerably by the
simulated tissue signal. Fig.~\ref{fig:unroll_corona_results}b and
Fig.~\ref{fig:unroll_corona_results}d illustrate the ground truth MIP images
of the UCA signal and tissue signal respectively, whereas
Fig.~\ref{fig:unroll_corona_results}c and
Fig.~\ref{fig:unroll_corona_results}e presents the MIP images of the
recovered UCA signal and tissue signal via CORONA.\@

Another approach that leverages deep algorithm unrolling is used for super resolution US~\cite{van2019deep,bar2021learned}, where UCAs are pinpointed and tracked through a sequence of US frames, to yield super-resolution images of the microvasculature, thus circumventing the diffraction limited resolution of conventional ultrasound. Unrolling is used to derive a robust method for super resolution US that is suited to various
imaging conditions and does not depend on prior knowledge such as the PSF of the system. 

By leveraging the signal structure, i.e., assuming that the UCAs distribution is sparse on a sufficiently high-resolution grid, and assuming that the US frames contain only UCAs signals (i.e., at absence of tissue clutter and
noise), the model has the same form as~\eqref{eq:unroll_basic_model}, where $\bx$ is a vector that describes the sparse microbubble distribution on a
high-resolution image grid, $\by$ is the vectorized image frame of the ultrasound sequence, $\bW$ is a dictionary matrix derived from the PSF of the system, and $\bn$ is a noise vector. Again ISTA can be employed to solve this
problem.

The unrolled network, named deep unrolled ultrasound localization microscopy
(ULM), is similar to LISTA, only that the soft-thresholding operator is
    substituted with a  sigmoid-based soft-thresholding operation to avoid
vanishing gradients during training~\cite{zhang2001thresholding}. The network
is trained using back-propagation under supervision of synthetically generated
2D images of point sources.

\begin{figure}
	\centering
	\includegraphics[width=\textwidth]{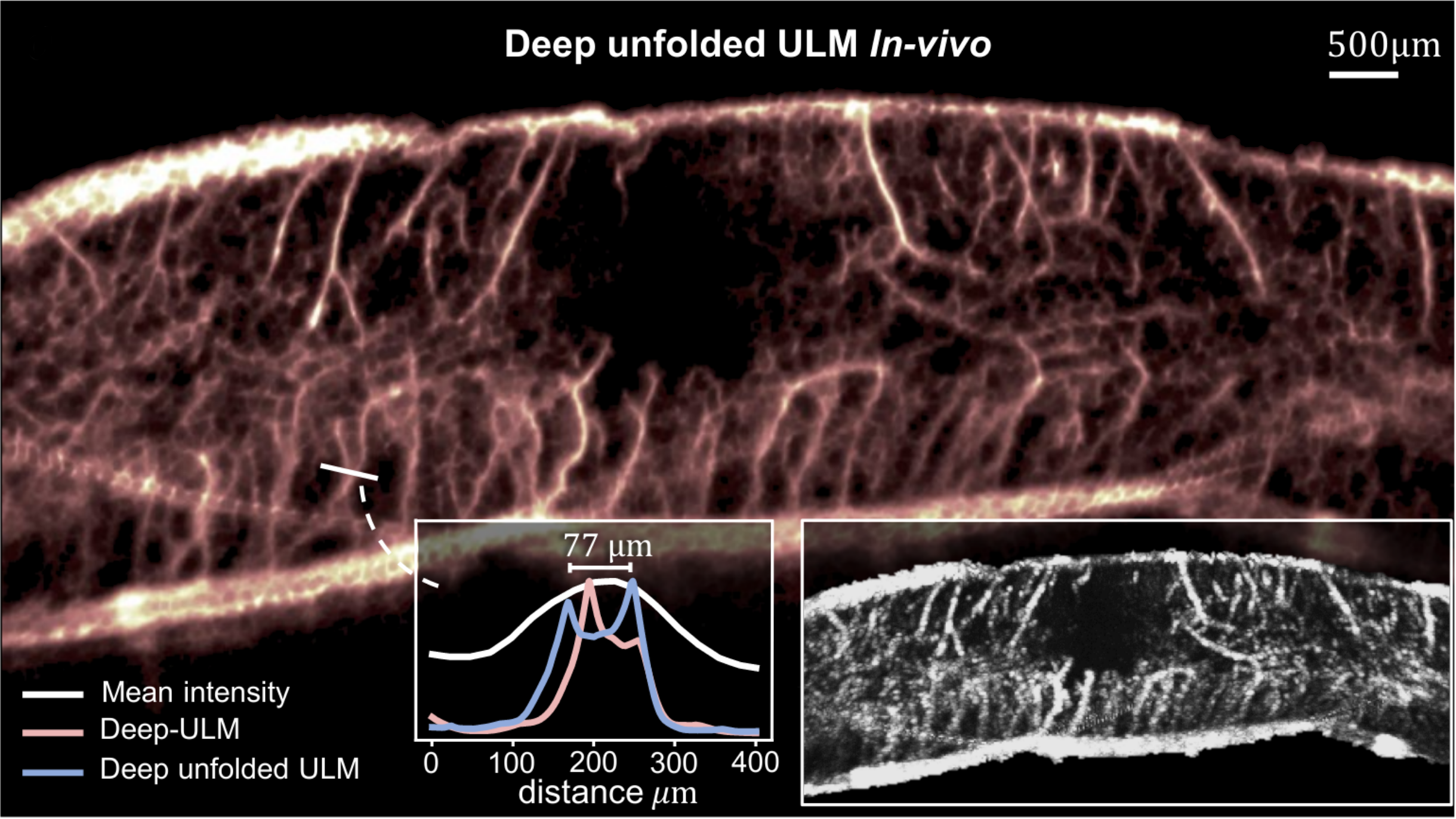}
	\caption{Deep unrolled ULM for super-resolution vascular imaging of a rat’s spinal cord. Right rectangle: standard maximum intensity projection across a sequence of frames. Left rectangle: spatial resolution comparison between the deep unrolled ULM, and Deep-ULM\@. Figure reproduced from~\protect\cite{van2019deep} with authors' permission.}\label{fig:unroll_ulm_results}
\end{figure}

\begin{figure}
	\centering
	\includegraphics[width=\textwidth]{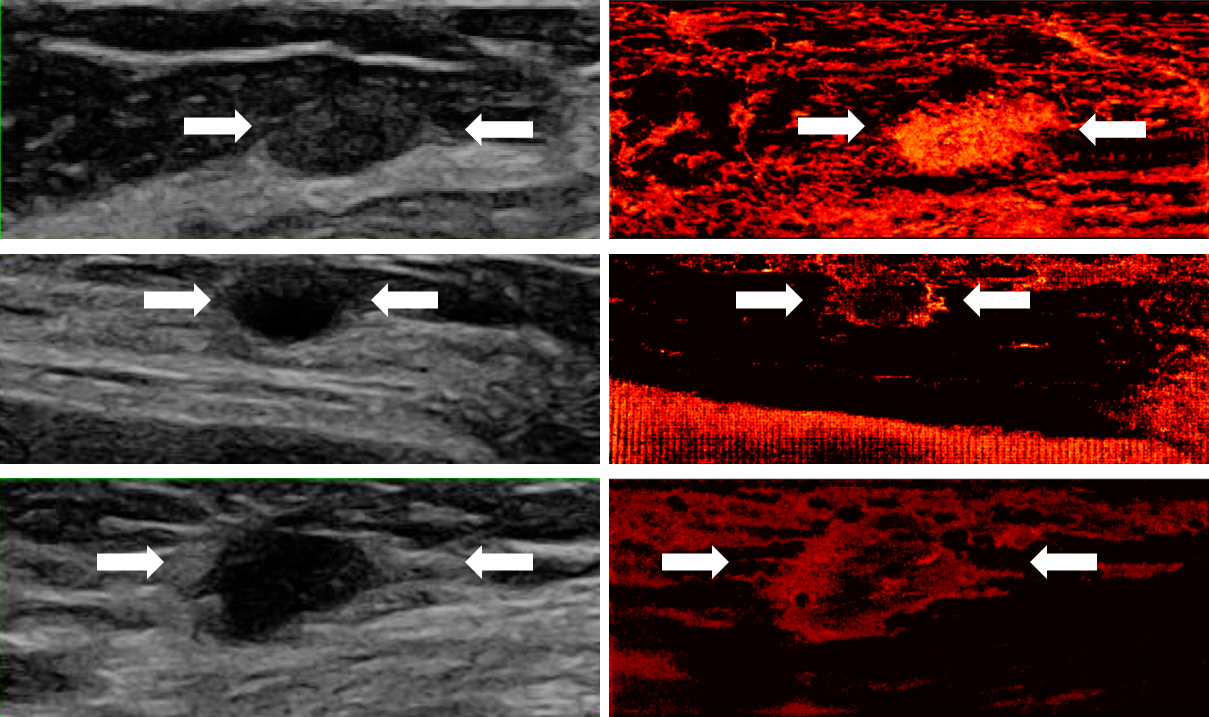}
	\caption{Super resolution demonstrations in human scans of three lesions in breasts of three patients. Left: B-mode images. Right: super resolution recoveries. The white arrows point at the lesions; Top: fibroadenoma (benign). The super resolution recovery shows an oval, well circumscribed mass with homogeneous high vascularization. Middle: cyst (benign). The super resolution recovery shows a round structure with high concentration of blood vessels at the periphery of the lesion. Bottom: invasive ductal carcinoma (malignant). The super resolution recovery shows an  irregular mass with ill-defined margins, high concentration of blood vessels at the periphery of the mass, and a low concentration of blood vessels at the center of the mass. Figure reproduced from~\protect\cite{bar2021learned} with authors' permission.}\label{fig:unroll_ulm_results_human}
\end{figure}

The authors trained a ten-layer deep network comprising 5 × 5 convolutional
kernels.  Fig.~\ref{fig:unroll_ulm_results} provides visual results of deep
unrolled ULM for super-resolution vascular imaging of a rat's spinal cord
exhibiting a smooth reconstruction of the vasculature. Compared with an
encoder–decoder approach devised for the same task and trained on the same
data, named deep-ULM as proposed in~\cite{van2019deep}, deep unrolled
ULM exhibited a drastically lower memory footprint and reduced power
consumption, in addition to achieving higher inference rates and improved
generalization when encountering new data. This can be seen in the spatial
resolution comparison in Fig.~\ref{fig:unroll_ulm_results}, showing the
superior capability of deep unrolled ULM to separate close structures. Recently, the method was harnessed to improve breast lesion characterization~\cite{bar2021learned}. Recoveries of three in vivo human scans of lesions in breasts of three patients were shown. Fig.~\ref{fig:unroll_ulm_results_human} presents the recoveries of the three lesions along with the B-Mode US images that are used in clinical practice. The results reveal vascular structures that are not seen in the B-Mode images. Furthermore, the recoveries show unique vascular structure for each lesion, thus assisting the differentiation between the lesions.

\subsection{Applications of Unrolling in Medical Resonance Imaging}\label{ssec:unroll_mri}
A typical Magnetic Resonance Imaging (MRI) system comprises the following
components: a primary magnet which exerts a primary magnetic field, gradient
magnets which create spatial variation of the magnetic field and allow spatial
encoding of the signal, Radio-Frequency (RF) coils which send out and receive a
series of RF pulses and magnetic field gradients, and a computer system which
decodes the received signal and reconstructs the original image of interest.
Formally speaking, the signal $\by(t)$ received in the receiver at time $t$ is
given by
\[
	\by(\bk,t)=\int \bx(x,y,z)\exp\{-j2\pi(k_{x}x+k_{y}y+k_{z}z)\}\mathrm{d}x\mathrm{d}y\mathrm{d}z,
\]
where $\bx$ is the image, $x,y,z$ is the spatial location, and $\bk=(k_x,k_y,k_z)$ is the $k$-space location at time $t$. In practice, a discretized model is commonly used:
\[
	\by(\bk_i)=\sum_l \bx(\br_l)\exp\{-j2\pi \bk_i\cdot\br_l\},
\]
where $\bk_i$ and $\br_l$ are $k$-space and spatial locations and $\cdot$
denotes the dot product between 3D vectors. That is, the measured signal $\by$
comprises a series of Fourier encoding of the signal $\bx$. In compressed MRI, only a
subset of Fourier samples are observed, which gives rise to the following
linear compressive imaging model:
\[
	\by=\bS\bF\bx,
\]
where $\bF\in\mathbb{C}^{m\times m}$ is the Fourier matrix and
$\bS\in{\{0,1\}}^{n\times m}$ is the subsampling matrix. In the MRI literature,
more advanced imaging models are developed to account for other factors such as
inhomogeneity of the magnetic fields, coupling between gradient coils, timing
inaccuracies and motion. For a thorough review of these models, see~\cite{doneva2020mathematical}.

To accelerate data acquisition of dynamic MRI, Qin {\it et
al.}~\cite{qin2018convolutional} propose a Convolutional Recurrent Neural
Network (CRNN) that captures temporal dependencies and reconstructs high
quality cardiac MR images from highly undersampled $k$-space data. They first
analyze the following classical regularized inverse problems:
\begin{equation}
	\min_{\bx}\cR(\bx)+\lambda\|\by-\bS\bF\bx\|_2^2,\label{eq:unroll_mri_obj}
\end{equation}
where $\bx$ comprises a sequence of complex-valued MR images, $\cR$ is the regularization functional and $\lambda$ is the (reciprocal)
regularization coefficient. By introducing an extra variable $\bz$,
problem~\eqref{eq:unroll_mri_obj} can be relaxed into:
\begin{equation}
	\min_{\bx,\bz}\lambda\|\by-\bS\bF\bx\|_2^2+\mu\|\bx-\bz\|_2^2+\cR(\bz),\label{eq:unroll_mri_relax}
\end{equation}
where $\mu>0$ is the penalty parameter. When $\mu\rightarrow\infty$, the
relaxed problem~\eqref{eq:unroll_mri_relax} reduces to the original
one~\eqref{eq:unroll_mri_obj}. Problem~\eqref{eq:unroll_mri_relax} can be
solved by alternately minimizing over $\bx$ and $\bz$:
\begin{align}
	\bz_{k+1}&\gets\arg\min_{\bz}\cR(\bz)+\mu\|\bx_k-\bz\|_2^2,\label{eq:unroll_mri_prox}\\
	\bx_{k+1}&\gets\arg\min_{\bx}\lambda\|\by-\bS\bF\bx\|_2^2+\mu\|\bx-\bz_{k+1}\|_2^2\nonumber\\
			 &=\bF^H\bm{\Lambda}\bF\bz_k+\frac{\lambda_0}{1+\lambda_0}\bS\bF^H\by\nonumber\\
			 &:=\cD\cC\left(\bz_k;\by,\lambda_0,\Omega\right),\label{eq:unroll_mri_dc}
\end{align}
where $\lambda_0=\frac{\lambda}{\mu}$, $\bm{\Lambda}$ is a diagonal matrix accounting
for the under-sampling scheme, and $\Omega$ is the index set of the acquired
$k$-space samples. Qin {\it et al.}~\cite{qin2018convolutional} propose to
replace~\eqref{eq:unroll_mri_prox} with a CRNN, providing the following
iterations:
\begin{align*}
	\bz_{k+1}&\gets\bx_k+\mathrm{CRNN}(\bx_{k}),\\
	\bx_{k+1}&\gets\cD\cC(\bz_{k+1};\by,\lambda_0,\Omega).
\end{align*}

The CRNN is a learnable RNN which integrates convolutions and evolves both
    over iterations and time. To this end, the authors introduce bidirectional
    convolutional recurrent units, evolving both over time and iterations, and
    convolutional recurrent units, evolving over iterations only. CRNN
    introduces inter-iteration connections so that information can be
    propagated across iterations, whereas in conventional unrolling techniques
    the iterations are typically disconnected. In addition, CRNN exploits
    temporal dependency and data redundancies of the dynamic MRI data by
    introducing temporal progressions. Different from typical RNNs, The weights
    for CRNN are independent across layers. The network is trained end-to-end
    by minimizing the MSE loss between the predicted MR images and ground truth
    data. Experimentally, CRNN achieves nearly 1dB improvement of PSNR for
    dynamic MRI, compared to state-of-the-art approaches including both
    iterative reconstruction and deep learning. In terms of running time, it is
also significantly faster than both classes of approaches.

In a different algorithmic front, Pramanik {\it et
al.}~\cite{pramanik2020deep} unroll the Structured Low-Rank (SLR) algorithm for
calibration-less parallel MRI and multishot MRI applications. The SLR algorithm
performs MR image reconstruction by lifting the signal into a structured matrix
and solving the following rank minimization problem:
\begin{align}
	\min_{\bm{\Gamma}}\quad&\mathrm{rank}\left[\cT\left(\cG\left(\widehat{\bm{\Gamma}}\right)\right)\right]\nonumber\\
	\text{subject to }\;&\bB=\cA(\bm{\Gamma})+\bP,\label{eq:unroll_low_rank}
\end{align}
where $\bm{\Gamma}$ is the matrix representing multi-channel images on
different coils, $\widehat{\cdot}$ denotes the Discrete Fourier Transform
(DFT), $\bB$ is the corresponding noisy under-sampled multi-channel Fourier
measurement, $\cA$ is the linear operator formed by composing the Fourier
transform with subsampling, and $\bP$ is the multi-channel noise
matrix. Here $\cG$ is the mapping from Fourier samples to their gradients, and $\cT$
is the lifting operator that lifts the signal into a higher dimensional
structured matrix. The low-rank constraint comes from the fact that
the matrix $\cT\left(\cG\left(\widehat{\bm{\Gamma}}\right)\right)$ has a high-dimensional
null space, which in turns originates from sparsity of edges in natural images.
For technical details, see~\cite{pramanik2020deep} and~\cite{ongie2017convex}.

Problem~\eqref{eq:unroll_low_rank} is difficult to tackle
directly due to its non-convex nature, and the following convex relaxation
usually serves as a tractable surrogate:
\begin{equation}
	\min_{\bm{\Gamma}}\,\|\cA(\bm{\Gamma})-\bB\|_2^2+\lambda\left\|\cT\left(\cG\left(\widehat{\bm{\Gamma}}\right)\right)\right\|_\ast,\label{eq:unroll_lr_relax}
\end{equation}
where $\lambda$ is the regularization coefficient and $\|\cdot\|_\ast$ is the
nuclear norm whose minimization promotes low-rank structures of the solution.
Problem~\eqref{eq:unroll_lr_relax} can be solved by the Iteratively Reweighted
Least Suqares (IRLS) algorithm which majorizes the nuclear norm with a weighted
Frobenuis norm and solves the problem:
\begin{equation}
	\min_{\bm{\Gamma},\bQ}\,\|\cA(\bm{\Gamma})-\bB\|_2^2+\lambda\left\|\cT\left(\cG\left(\widehat{\bm{\Gamma}}\right)\right)\bQ\right\|_F^2.\label{eq:unroll_lr_reweight}
\end{equation}

\begin{figure*}
	\includegraphics[width=\textwidth]{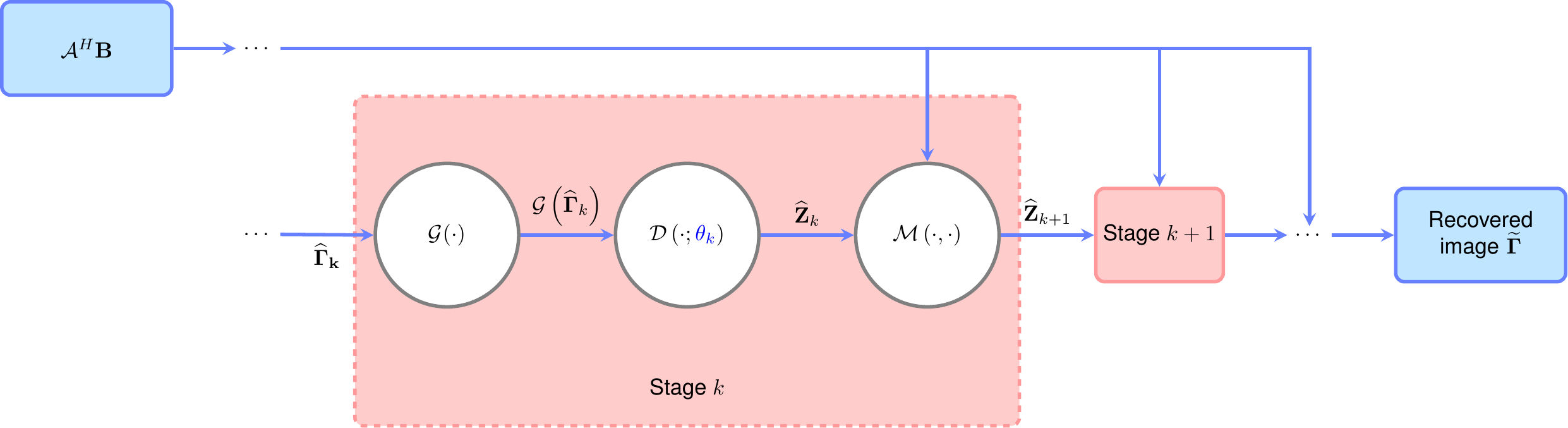}
	\caption{Diagram representation of DSLR~\protect\cite{pramanik2020deep}: at each stage, the input $\bm{\Gamma}_k$ passes through a gradient operator, a CNN defined in~\eqref{eq:unroll_lr_cnn}, and a data consistency layer defined by~\eqref{eq:unroll_lr_dc}. Trainable parameters are colored in blue.}\label{fig:unroll_deepslr}
\end{figure*}

By introducing an auxiliary variable $\widehat{\bZ}$,
problem~\eqref{eq:unroll_lr_reweight} can be further relaxed into:
\begin{equation}
	\min_{\bm{\Gamma},\bQ,\widehat{\bZ}}\,\|\cA(\bm{\Gamma})-\bB\|_2^2+\lambda\left\|\cT\left(\widehat{\bZ}\right)\bQ\right\|_F^2+\beta\left\|\cG\left(\widehat{\bm{\Gamma}}\right)-\widehat{\bZ}\right\|_2^2,\label{eq:unroll_irls_relax}
\end{equation}
where $\beta>0$ is the penalty coefficient. Problem~\eqref{eq:unroll_irls_relax} can be solved by alternately minimizing over $\bm{\Gamma}$, $\widehat{\bZ}$ and $\bQ$:
\begin{align}
	\bm{\Gamma}_{k+1}&\gets\arg\min_{\bm{\Gamma}}\,\|\cA(\bm{\Gamma})-\bB\|_2^2+\beta\left\|\cG\left(\widehat{\bm{\Gamma}}\right)-\widehat{\bZ}_k\right\|_2^2\nonumber\\
					 &={\left(\cA^H\cA+\beta\cG^H\cG\right)}^{-1}\left(\cA^H\bB+\beta\cG^H\left(\widehat{\bm{\bZ}_k}\right)\right)\nonumber\\
					 &:=\cM\left(\cA^H\bB,\bZ_k\right)\label{eq:unroll_lr_dc}\\
	\widehat{\bm{\bZ}}_{k+1}&\gets\arg\min_{\widehat{\bZ}}\,\beta\left\|\cG\left(\widehat{\Gamma}_{k+1}\right)-\widehat{\bZ}\right\|_2^2+\lambda\left\|\cT\left(\widehat{\bm{Z}}\right)\bQ\right\|_F^2\nonumber\\
							&={\left[\bI+\frac{\lambda}{\beta}{\cJ(\bQ_k)}^H\cJ(\bQ_k)\right]}^{-1}\cG\left(\widehat{\bm{\Gamma}}_{k+1}\right),\label{eq:unroll_lr_prox}\\
	\bQ_{k+1}&={\left[{\cT\left(\cG\left(\widehat{\bm{\Gamma}}_{k+1}\right)\right)}^H\cT\left(\cG\left(\widehat{\bm{\Gamma}}_{k+1}\right)\right)+\epsilon_{k+1}\bI\right]}^{-\frac{1}{4}},\nonumber
\end{align}
where $\cJ$ is a lifting operator such that $\cJ(\bQ)$ represents a filter bank. Pramanik {\it et al.}~\cite{pramanik2020deep} propose to replace the update step~\eqref{eq:unroll_lr_prox} with a denoising CNN, and thus adopt the following iteration procedures:
\begin{align}
	\widehat{\bZ}_k&\gets\cD\left[\cG\left(\widehat{\bm{\Gamma}}_k\right);\theta_k\right],\label{eq:unroll_lr_cnn}\\
	\widehat{\bm{\Gamma}}_{k+1}&\gets\cM\left(\cA^H\bB,\bZ_k\right),\nonumber
\end{align}
where $\cD\left(\cdot;{\theta_k}\right)$ is a CNN with parameter $\theta_k$.
The unrolled network, called Deep SLR (DSLR), is illustrated visually
in Fig.~\ref{fig:unroll_deepslr}. At implementation, DSLR adopts a hybrid
regularization scheme, which incorporates an additional prior to exploit
image-domain redundancies. Training is performed using the MSE loss.

\begin{figure*}
	\centering
	\includegraphics[width=\textwidth]{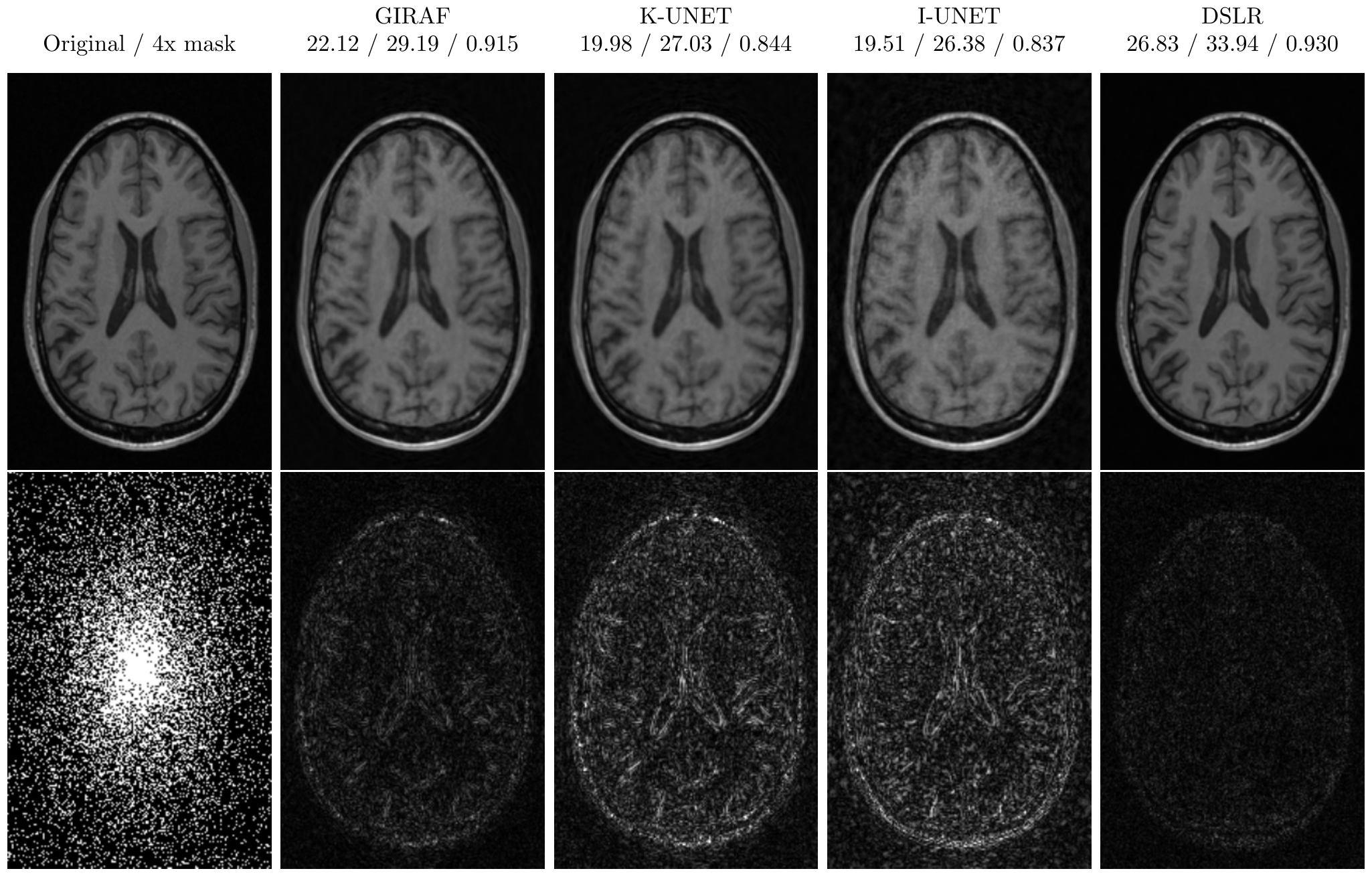}
	\caption{Experimental results on reconstruction of 4x accelerated single-channel brain data. The data was under-sampled using a Cartesian 2D non-uniform variable-density mask. The SNR (dB) / PSNR (dB) / SSIM values are also included for quantitative comparisons. The top row and the bottom row show reconstructed images (in magnitude) and corresponding error images, respectively. DSLR is proposed in~\protect\cite{pramanik2020deep}, while the SLR algorithm (GIRAF)~\protect\cite{ongie2017fast}, $k$-space UNet (K-UNET)~\protect\cite{han2019k} and image domain UNET (I-UNET) are state-of-the-art techniques for comparison. Images are courtesy of Matthews Jacob at University of Iowa~\protect\cite{pramanik2020deep}.}\label{fig:unroll_deepslr_experiments}
\end{figure*}

Experimentally, DSLR demonstrates superior performance over state-of-the-art
techniques on several evaluation metrics: SNR, PSNR, and SSIM\@. In addition, DSLR
runs significantly faster compared with SLR methods as it is free from
expensive operations such as Singular Value Decomposition (SVD), bringing down
the running time from tens of minutes to less than a second.
Figure~\ref{fig:unroll_deepslr_experiments} provides a visual comparison
between DSLR and state-of-the-art MRI algorithms, where it is clearly
observed that DSLR achieves much more accurate recovery with lower errors and
higher quantitative scores.

\subsection{Unrolling Techniques across Multiple Biomedical Imaging Modalities}\label{ssec:unroll_generic}
In addition to application-specific approaches, there are numerous techniques
applicable to multiple medical imaging modalities. Different medical
imaging problems may share a similar forward model and have common prior
structure, which enables the development of an abstract imaging algorithm that
can be adapted towards each particular problem. An example that unrolls
the well-known Alternating Direction Method of Multipliers (ADMM) algorithm is
the ADMM-CSNet proposed in~\cite{yang2018admm}, where Yang {\it et al.} develop
a Compressive Sensing (CS) technique for a linear imaging model. Specifically,
the measurement vector $\by\in\mathbb{C}^m$ is collected through
$\by\approx\bW\bx$ where $\bW\in\mathbb{C}^{n\times m}$ is a measurement matrix
with $m>n$ and $\bx\in\mathbb{R}^m$ is the signal to be recovered.

The CS problem is typically solved by exploiting the underlying sparse
structure of the signal $\bx$ in a certain transformation
domain~\cite{eldar_2012_compressed}. In~\cite{yang2018admm}, a slightly generalized CS model is employed, which amounts to solving the following optimization problem~\cite{yang2018admm}:
\begin{align}
	\min_{\bx}\frac{1}{2}\|\bW\bx-\by\|_2^2+\sum_{i=1}^C\lambda_i g(\bD_i\bx),\label{eq:unroll_general_cs}
\end{align}
where $\lambda_i$'s are positive regularization coefficients, $g(\cdot)$ is a sparsity-inducing function, and ${\{\bD_i\}}_{i=1}^C$ is a sequence of $C$ operators, which effectively perform linear filtering operations. Concretely, $\bD_i$ can be taken as a wavelet transform and $g$ can be chosen as the $\ell^1$ norm. However, instead of resorting to analytical filters and prior functions, in~\cite{yang2018admm} they are learned from training data through unrolling.

\begin{figure*}[h!]
	\includegraphics[width=\textwidth]{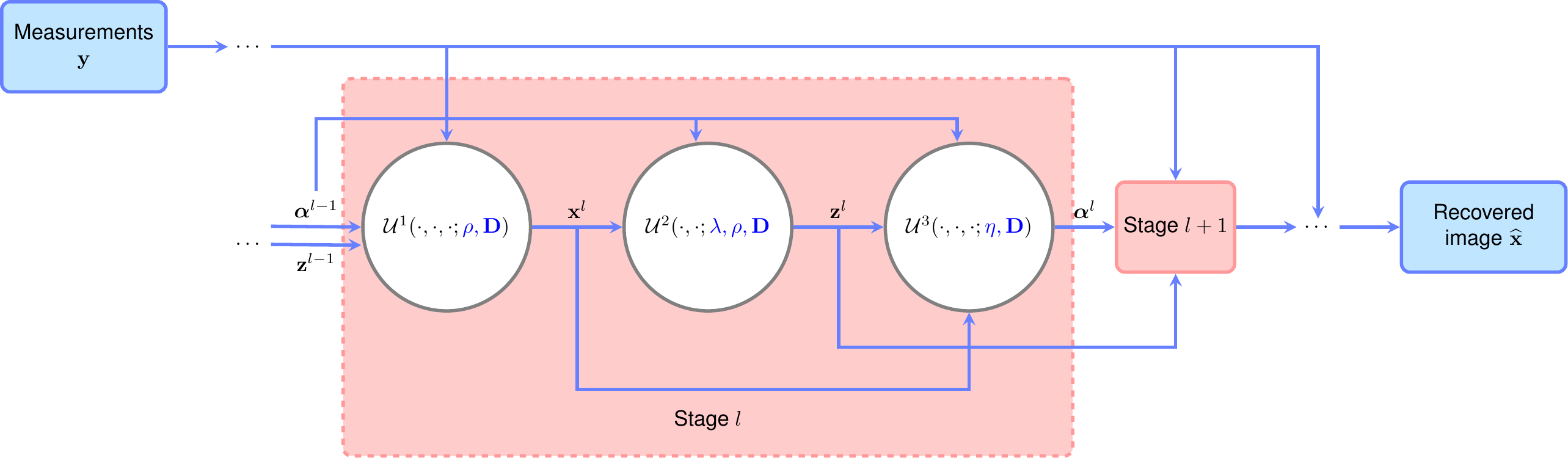}
	\caption{Diagram representation of ADMM-CSNet~\protect\cite{yang2018admm}: each stage comprises a series of inter-related operations, whose analytic forms are given in~\eqref{eq:admm_csnet_update}. The trainable parameters are colored in blue.}\label{fig:unroll_admm_csnet}
\end{figure*}

In general, there are numerous techniques for solving~\eqref{eq:unroll_general_cs} such as the ISTA algorithm discussed in~\ref{ssec:unroll_ista}. Among them a simple yet efficient algorithm is ADMM~\cite{boyd_2011_distributed}, which has found wide applications in various imaging domains. To employ it, problem~\eqref{eq:unroll_general_cs} is first recast into a constrained minimization problem through variable splitting, by introducing variables ${\{\bz_i\}}_{i=1}^C$:
\begin{align}
	\min_{\bx,{\{\bz\}}_{i=1}^C}&\frac{1}{2}\|\bW\bx-\by\|_2^2+\sum_{i=1}^C\lambda_i g(\bz_i),\nonumber\\
	\text{subject to }&\bz_i=\bD_i\bx,\quad\forall i.\label{eq:split_cs}
\end{align}
The corresponding augmented Lagrangian is then formed as:
\begin{align}
	\cL_\rho(\bx,\bz;\balpha_i)&=\frac{1}{2}\|\bW\bx-\by\|_2^2+\sum_{i=1}^C\lambda_i g(\bz_i)+\frac{\rho_i}{2}\|\bD_i\bx-\bz_i+\balpha_i\|_2^2,\label{eq:aug_lagrangian}
\end{align}
where ${\{\balpha_i\}}_{i=1}^C$ are the dual variables and ${\{\rho_i\}}_{i=1}^C$ are positive penalty coefficients. ADMM then alternately minimizes~\eqref{eq:aug_lagrangian} followed by a dual variable update, leading to the following iterations:
\begin{align}
	\bx^l&={\left(\bW^H\bW+\sum_{i=1}^C\rho_i\bD_i^T\bD_i\right)}^{-1}\bigl[\bW^H\by+\sum_{i=1}^C \rho_i {\bD_i}^T\left({\bz_i}^{l-1}-{\balpha_i}^{l-1}\right)\bigr]\nonumber\\
		 &:=\cU^1\left\{\by,\balpha_i^{l-1},\bz_i^{l-1};\rho_i,\bD_i\right\},\nonumber\\
		\bz_i^l&=\cP_{g}\left\{\bD_i\bx^l+\balpha_i^{l-1};\frac{\lambda_i}{\rho_i}\right\}\label{eq:admm_csnet_update}\\
			   &:=\cU^2\left\{\balpha_i^{l-1},\bx^l;\lambda_i,\rho_i,\bD_i\right\},\nonumber\\
		\balpha_i^l&=\balpha_i^{l-1}+\eta_i(\bD_i\bx^l-\bz^l_i)\nonumber\\
				   &:=\cU^3\left\{\balpha_i^{l-1},\bx^l,\bz_i^l;\eta_i,\bD_i\right\},\quad\forall i,\nonumber
\end{align}
where $\eta_i$'s are constant parameters, and $\cP_g\{\cdot;\lambda\}$ is the proximal mapping for $g$ with parameter $\lambda$. The unrolled network can thus be constructed by concatenating these operations. The parameters $\lambda_i,\rho_i,\eta_i,\bD_i$ in each layer will be learned from real datasets. Fig.~\ref{fig:unroll_admm_csnet} depicts the resulting network architecture. In~\cite{yang2018admm} the authors discuss several implementation issues, including efficient matrix inversion and the analytic back-propagation rules. The network is trained by minimizing a normalized version of the Root-Mean-Square-Error (RMSE).

ADMM-CSNet demonstrates its efficacy through various experiments. In
particular, for MRI applications, ADMM-CSNet achieves the same reconstruction
accuracy using $10\%$ less sampled data and speeds up recovery by around $40$
times compared to conventional iterative methods. When compared with
state-of-the art deep networks, it exceeds their performance by a margin of
around 3dB PSNR under $20\%$ sampling rate on brain data. 

In a similar fashion, Aggrawal {\it et al.}~\cite{aggarwal2019modl} unrolls an
alternating minimization algorithm over a regularized image reconstruction
problem. By integrating a CNN into the image prior, they start with the
following optimization problem:
\begin{equation}
	\min_{\bx}\|\bW\bx-\by\|_2^2+\lambda\left\|\bx-\cD_\theta(\bx)\right\|_2^2,\label{eq:unroll_modl_obj}
\end{equation}
where $\bx\in\mathbb{C}^m$ is the underlying image to be recovered,
$\by\in\mathbb{C}^n$ is the measurement vector, $\bW\in\mathbb{C}^{n\times m}$
is the measurement model matrix, and $\cD_\theta$ is a denoising CNN carrying
parameters $\theta$. Aggrawal {\it et al.} propose to solve~\eqref{eq:unroll_modl_obj} via the following iterations:
\begin{align}
	\bz_k&\gets\cD_\theta(\bx_k),\nonumber\\
	\bx_{k+1}&\gets\arg\min_{\bx}\|\bW\bx-\by\|_2^2+\lambda\|\bx-\bz_k\|_2^2\nonumber\\
			 &={\left(\bW^H\bW+\lambda\bI\right)}^{-1}\left(\bW^H\by+\lambda\bz_k\right),\label{eq:unroll_conj_grad}
\end{align}
where $\bI\in\mathbb{R}^{m\times m}$ is the identity matrix. The
network parameters $\theta$ are shared across different iterations. The
unrolled network, dubbed MoDL, has a very similar structure to DSLR
in Fig.~\ref{fig:unroll_deepslr}. Training of MoDL is conducted by minimizing the MSE loss.

For applications such as MRI, $\bW$ corresponds to a composition of the
sampling operator with the Fourier matrix and thus the iteration
step~\eqref{eq:unroll_conj_grad} admits a simple analytic formula. However, in more complex cases, such as multichannel MRI,
 the operator $\bW^H\bW+\lambda\bI$ is not analytically invertible and
iterative numerical methods such as Conjugate Gradient (CG) need to be plugged
into the network. Aggrawal {\it et al.}~\cite{aggarwal2019modl} verify that,
although CG is an iterative technique, the intermediate results need not be
stored to perform back-propagation. Instead, another CG step simply needs to be
applied in the back-propagation step.

Numerous experiments verify the effectiveness of MoDL\@. When compared to other
deep learning frameworks, MoDL achieves nearly 3dB higher PSNR values for 6x acceleration on their own dataset acquired through 3D T2 CUBE.\@ Furthermore, as MoDL incorporates the forward model $\by=\bW\bx$, it is
relatively insensitive to the undersampling patterns in MRI\@. In particular,
when trained on 10x acceleration setting, MoDL is capable of faithfully
recovering the images under 12x and 14x acceleration settings.

\section{Perspectives and Recent Trends}\label{sec:unroll_trends}
We have so far reviewed several successes of algorithm unrolling, including
both conceptual and practical breakthroughs in many biomedical imaging topics.
In this section we will put algorithm unrolling into perspective, in order to
understand why it is beneficial compared with traditional iterative algorithms
and contemporary DNNs. Furthermore, we summarize some recent trends that
we observe in many biomedical imaging works.

\subsection{Why is Unrolling So Effective for Biomedical Imaging?}\label{ssec:unroll_distill}

In recent years, algorithm unrolling has proved highly effective in achieving
superior performance and higher efficiency in many practical domains, while
inheriting interpretability from the underlying iterative algorithms. In
addition, it frequently demonstrates improved robustness under deviations of
training data and forward models compared with generic DNNs. A question that
naturally arises is, why is it so powerful?

From a machine learning perspective, a fundamental trade-off for learning-based
models is the \emph{bias-variance trade-off}~\cite{friedman2001elements}. Models that have low capacity,
such as linear models, typically generalize better but may prove inadequate to
capture complicated data patterns. Such models generally exhibit high bias and
low variance. In contrast, models that have abundant capacity, such as DNNs,
are capable of fitting sophisticated functions but have high risk of
overfitting.

\begin{figure*}[h]
	\centering
	\includegraphics[width=0.7\textwidth]{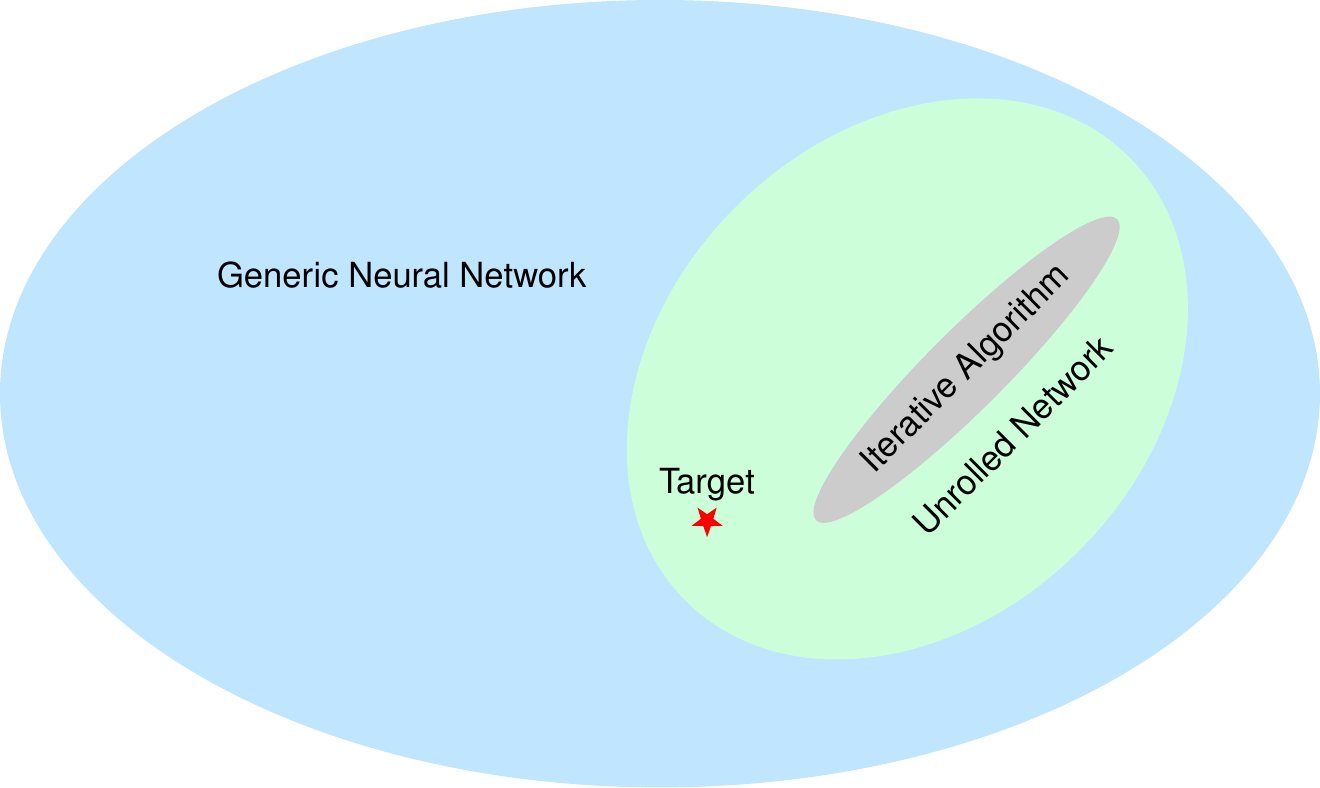}
	\caption{A high-level unified interpretation of algorithm unrolling from a functional approximation perspective: the ellipses depict the scope of functions that can be approximated by each category of methods. Compared with iterative algorithms which have limited representation power and usually underfit the target, unrolled networks often better approximate the target thanks to their higher representation power. On the other hand, unrolled networks have lower representation power than generic neural networks but usually generalize better in practice, hence providing an attractive balance.}\label{fig:unify}
\end{figure*}

Unrolled networks achieve a more favorable
bias-variance trade-off compared with alternative techniques~\cite{yang2018admm}, in particular
generic neural networks and iterative algorithms. Fig.~\ref{fig:unify} provides
a high-level illustration of the representation capability of these classes of
techniques, from a functional approximation perspective. The area of various class of methods represents their capability in fitting functions, i.e., their modeling capacity as functional approximators. A
traditional iterative algorithm spans a relatively small subset of the
functions compared to deep learning techniques, and thus has low variance but high bias. Indeed, in practice iterative algorithms typically underperform deep learning techniques which provides as a strong evidence of their low modeling capacity. Therefore, there is
generally an irreducible gap between its spanned set and the target function.
Nevertheless, in implementing iterative methods, typically the user tweaks the architecture
and fine-tunes parameters, rendering it capable of approximating a given
target function reasonably well and effectively reducing the gap. On the other
hand, iterative algorithms generalize relatively well in limited training
scenarios thanks to their low variance. In biomedical imaging, the data is
generally difficult to gather due to high costs of imaging devices and
requirements of patients. Therefore, iterative algorithms have played a critical
role for many years.

On the other hand, a generic neural network is capable of more accurately
approximating the target function thanks to its universal approximation
capability~\cite{sonoda2017neural}.  As shown in Fig.~\ref{fig:unify}, it
constitutes a large subset in the function space. In practice, it has been
    empirically supported that DNNs can approximate complicated mappings as
long as the data is sufficient. Nowadays, DNNs typically consists of an
enormous number of parameters and have low bias but high variance. The high
dimensionality of parameters requires abundant training samples, otherwise
generalization becomes a serious issue. Furthermore, searching for the
target function out of its huge spanned space is rather difficult which
poses great challenges in training.

In contrast, the unrolled networks are typically constructed by expanding the
capacity of iterative algorithms, through generalizations of the underlying
iterations. Therefore, it integrates domain knowledge from iterative
algorithms and can approximate the target function more accurately, while
spanning a relatively small subset in the function space. Reduced size of the
search space alleviates the burden of training and requirement of large scale
training datasets. In addition, unrolled networks have better generalizability
thanks to their lower variance and are less prone to overfitting. As an
intermediate model between generic networks and iterative algorithms, unrolled
networks typically have relatively low bias and variance simultaneously.

\subsection{ Emerging Unrolling Trends for Biomedical Imaging}\label{ssec:unroll_implementation}
As we have seen through various case studies, there are a few common trends
shared by many methods, which are summarized below.

\subsubsection{Using DNNs as learnable operators}
As discussed in Section~\ref{sec:unroll_development}, an important procedure in
constructing unrolled networks is to generalize the functional form per
iteration. A commonly employed technique is to substitute certain operators
with DNNs. In particular, proximal operators which manifest in various
regularized inverse problems are frequently replaced by parameterized deep
networks such as CNNs~\cite{gupta2018cnn,hauptmann2018model,adler2018learned}
or RNNs~\cite{qin2018convolutional}. This substitution technique is reminiscent
of the Plug-and-Play scheme~\cite{ahmad2020plug}, where the proximal operators
are replaced with an off-the-shelf image denoiser such as Block-Matching and 3D
filtering (BM3D)~\cite{dabov2007image}.

\subsubsection{Exploring broader classes of algorithms}
In the early stage of unrolling research the focus was primarily in
unrolling ISTA-like algorithms~\cite{gregor_learning_2010,jin_deep_2017}.
Nowadays,  researchers are pursuing other alternatives such as
PGD~\cite{gupta2018cnn}, ADMM~\cite{yang2018admm},
and IRLS~\cite{pramanik2020deep}, to name a few. This trend has contributed greatly
to the novelty and variety of unrolling approaches.

\subsubsection{Balancing performance and efficiency trade-offs}
In practice, when designing unrolled deep networks, there is a fundamental
trade-off between performance and efficiency: wider and deeper networks with
numerous parameters generally achieve better performance but are usually less
efficient and more data demanding. Therefore, under practical constraints
important design choices have to be made. One distinction is whether to use
shared parameters across all the layers or adopt layer-specific parameters.
Another issue is how to determine the depth of the networks, i.e., number of
iterations in the algorithm.

\section{Conclusions}\label{sec:unroll_conclusions}
In this chapter we discussed algorithm unrolling, a systematic approach to
construct deep networks out of iterative algorithms, and its applications to
biomedical imaging. We first provided a comprehensive tutorial on algorithm
unrolling and how to leverage it to construct deep networks. We then
showcased how algorithm unrolling can be applied in biomedical imaging by
discussing concrete examples in different medical imaging modalities. We next put
algorithm unrolling into perspective, illustrated why it can be so powerful from
a bias-variance trade-off standpoint, and summarized several recent trends in
unrolling research.

To facilitate future research, we conclude this chapter by briefly discussing
open challenges and suggesting possible future research directions. First,
although algorithm unrolling has inherited the merits of iterative algorithms
to a large extent, customizations of the iteration procedures might undermine
some of them. In particular, convergence guarantees may no longer hold.
Therefore, it is interesting to extend the theories around iterative algorithms
to incorporate interesting unrolled networks. Second, as algorithm unrolling
serves as a valuable bridge between iterative algorithms and deep networks, it
can be regarded as a tool to understand why deep networks are so effective in
practical imaging applications, in order to complement their lack of
interpretability. Finally, in biomedical imaging training data is relatively
deficient. Since algorithm unrolling has already demonstrated superior
generalizability in many previous works, it is interesting to exploit it as a
viable alternative under limited training scenarios, in addition to
 and/or integrated with semi-supervised learning.

\bibliography{unrolling}\label{refs}

\end{document}